\documentclass[twoside,11pt]{article}
\usepackage{jmlr2e}

\usepackage{amsmath, amssymb} 
\usepackage{graphicx, graphics, epsfig, color}
\usepackage{adjustbox}
\usepackage{subfigure}
\usepackage{tikz, pgfplots}
\usepackage{algorithm, algorithmic}
\usepackage{float}
\usepackage{multirow}
\usepackage{cuted, flushend}
\usepackage{midfloat}
\usepackage{bm}
\usepackage{fancyhdr}
\usepackage{enumerate}
\usepackage{siunitx}
\usepackage{ulem}
\pgfplotsset{compat=1.4}
\usetikzlibrary{backgrounds,spy}
\pgfdeclarelayer{background}
\pgfsetlayers{background,main}

\tikzset{new spy style/.style={spy scope={%
        magnification=5,
        size=1.25cm,
        connect spies,
        every spy on node/.style={
                rectangle,
                draw,
        },
        every spy in node/.style={
                draw,
                rectangle,
                fill=gray!40,
        }
}}}

\newcounter{cassumption}
\newtheorem{assumption}[cassumption]{Assumption}

\def\RR{\mathbb{R}}

\def\trans{{\sf T}}

\def\asto{ {\overset{\rm a.s.}{\longrightarrow}} }

\def\ftau{f(\tau)}

\def\fftau{f'(\tau)}

\def\ffftau{f''(\tau)}

\def\muctra{\mu_a^{\circ}}
\def\muctrb{\mu_b^{\circ}}

\DeclareMathOperator{\tr}{{\rm tr}}
\DeclareMathOperator{\argmin}{{\rm argmin}}
\DeclareMathOperator{\argmax}{{\rm argmax}}

\author{Xiaoyi Mai, Romain Couillet}

\jmlrheading{XX}{2017}{X-X}{XX/XX; Revised XX/XX}{XX/XX}{Mai, Couillet}

\ShortHeadings{A random matrix analysis of semi-supervised learning}{Mai, Couillet}

\firstpageno{1}

\begin{document}

\title{A random matrix analysis and improvement of semi-supervised learning for large dimensional data}

\author{\name Xiaoyi Mai \email xiaoyi.mai@l2s.centralesupelec.fr \\ \addr CentraleSup\'elec -- LSS -- Universit\'e ParisSud (Gif-sur-Yvette, France).
	\AND
	\name Romain Couillet \email romain.couillet@centralesupelec.fr \\ \addr CentraleSup\'elec -- LSS -- Universit\'e ParisSud (Gif-sur-Yvette, France).
        }

\editor{XX XX}

\maketitle

%
\begin{abstract}%
This article provides an original understanding of the behavior of a class of graph-oriented semi-supervised learning algorithms in the limit of large and numerous data. It is demonstrated that the intuition at the root of these methods collapses in this limit and that, as a result, most of them become inconsistent. Corrective measures and a new data-driven parametrization scheme are proposed along with a theoretical analysis of the asymptotic performances of the resulting approach. A surprisingly close behavior between theoretical performances on Gaussian mixture models and on real datasets is also illustrated throughout the article, thereby suggesting the importance of the proposed analysis for dealing with practical data. As a result, significant performance gains are observed on practical data classification using the proposed parametrization.
\end{abstract}
%
%

\section{Introduction}
\label{sec:intro}
Semi-supervised learning consists in classification schemes combining few labelled and numerous unlabelled data. With the advent of the big data paradigm, where supervised learning implies the impossible pre-labelling of sometimes millions of samples, these so-far marginal methods are attracting a renewed attention. Its appeal also draws on its providing an alternative to unsupervised learning which excludes the possibility to exploit known data. We refer to \citep{chapelle2006ssl} for an overview.

An important subset of semi-supervised learning methods concerns graph-based approaches. In these, one considers data instances $x_{1},\cdots,x_{n}\in\mathbb{R}^{p}$ as vertices on a graph with edge weights $W_{ij}$ encoding their similarity, which is usually defined through a kernel function $f$, as with radial kernels of the type $W_{ij}=f(\|x_{i}-x_{j}\|^{2}/p)$ which we shall focus on in this article. The motivation follows from one's expectation that two instances with a strong edge weight tend to belong to the same class and thus vertices of a common class tend to aggregate. Standard methods for recovering the classes of the unlabelled data then consist in various random walk \citep{jaakkola2002partially} or label propagation \citep{zhu2002learning} algorithms on the graph which softly allocate ``scores'' for each node to belong to a particular class. These scores are then compared for each class in order to obtain a hard decision on the individual unlabelled node class. A popular, and widely recognized as highly performing, example is the PageRank approach \citep{avrachenkov2011generalized}.

Many of these algorithms also have the particularity of having a closed-form and quite interrelated expression for their stationary points. These stationary points are also often found to coincide with the solutions to optimization problems under constraints, independently established. 
This is notably the case of \citep{zhu2003semi} under equality constraints for the labelled nodes or of \citep{belkin2004regularization, delalleau2005efficient} where a relaxation approach is used instead to allow for modifications of the value of labelled nodes -- this ensuring that erroneously labelled data or poorly informative labelled data do not hinder the algorithm performance. As is often the case in graph-related optimization, a proper choice of the matrix representative of the inter-data affinity is at the core of scientific research and debates and mainly defines the differences between any two schemes. In particular, \citep{joachims2003transductive} suggests the use of a standard Laplacian representative, where \citep{zhou2004learning} advises for a normalized Laplacian approach. These individual choices correspondingly lead to different versions of the label propagation methods on the graph, as discussed in \citep{avrachenkov2011generalized}. 

\medskip

A likely key reason for the open-ended question of a most natural choice for the graph representative arises from these methods being essentially built upon intuitive reasoning arising from low dimensional data considerations rather than from mostly inaccessible theoretical results. Indeed, the non-linear expression of the affinity matrix $W$ as well as the rather involved form assumed by the algorithm output (although explicit) hinder the possibility to statistically evaluate the algorithm performances for all finite $n,p$, even for simple data assumptions. The present article is placed instead under a large dimensional data assumption, thus appropriate to the present bigdata paradigm, and proposes instead to derive, for the first time to the best of the authors' knowledge, theoretical results on the performance of the aforementioned algorithms in the large $n,p$ limit for a certain class of statistically distributed data $x_1,\ldots,x_n\in\mathbb{R}^p$. Precisely due to the large data assumption, as we shall observe, most of the intuition leading up to the aforementioned algorithms collapse as $n,p\to\infty$ at a similar rate, and we shall prove that few algorithms remain consistent in this regime. Besides, simulations on not-so-large data (here the $p=784$ MNIST images \citep{lecun1998mnist} will be shown to strongly adhere the predicted asymptotic behavior.

Specifically, recall that the idea behind graph-based semi-supervised learning is to exploit the similarity between data points and thus expect a clustering behavior of close-by data nodes. In the large data assumption (i.e., $p\gg1$), this similarity-based approach suffers a curse of dimensionality. As the span of $\mathbb{R}^{p}$ grows exponentially with the data dimension $p$, when $p$ is large, the data points $x_{i}$ (if not too structured) are in general so sparsely distributed that their pairwise distances tend to be similar regardless of their belonging to the same class or not. The Gaussian mixture model that we define in Subsection~\ref{sec:model} and will work on is a telling example of this phenomenon; as we show, in a regime where the classes ought to be separable (even by unsupervised methods \citep{couillet2015kernel}), the normalized distance $\|x_{i}-x_{j}\|/\sqrt{p}$ of two random different data instances $x_{i}$ and $x_{j}$ generated from this model converges to a constant \textit{irrespective of the class of $x_{i}$ and $x_{j}$} in the Gaussian mixture and, consequently, the similarity defined by $W_{ij}=f(\|x_{i}-x_{j}\|^{2}/p)$ is asymptotically the same for all pairs of data instances. This behavior should therefore invalidate the intuition behind semi-supervised classification, hence likely render graph-based methods ineffective. Nonetheless, we will show that sensible classification on datasets generated from this model can still be achieved provided that appropriate amendments to the classification algorithms are enforced.

\medskip

Inspired by \citep{avrachenkov2011generalized}, we generalize here the algorithm proposed in \citep{zhu2003semi} by introducing a normalization parameter $\alpha$ in the cost function in order to design a large class of regularized affinity-based methods, among which are found the traditional Laplacian- and normalized Laplacian-based algorithms. The generalized optimization framework is presented in Section~\ref{sec:optimization}. 

The main contribution of the present work is to provide a quantitative performance study of the generalized graph-based semi-supervised algorithm for large dimensional Gaussian-mixture data and radial kernels, technically following the random matrix approach developed in \citep{couillet2015kernel}. Our main findings are summarized as follows:
\begin{itemize}
\item Irrespective of the choice of the data affinity matrix, the classification outcome is strongly biased by \textit{the number of labelled data from each class} and unlabelled data tend to be classified into the class with most labelled nodes; we propose a normalization update of the standard algorithms to correct this limitation.
\item Once the aforementioned bias corrected, the choice of the affinity matrix (and thus of the parameter $\alpha$) strongly impacts the performances; most importantly, within our framework, both \textit{standard Laplacian} ($\alpha=0$ here) and \textit{normalized Laplacian-based} ($\alpha=-\frac12$) methods, although widely discussed in the literature, fail in the large dimensional data regime. Of the family of algorithms discussed above, only the \textit{PageRank} approach ($\alpha=-1$) is shown to provide asymptotically acceptable results.
\item The scores of belonging to each class attributed to individual nodes by the algorithms are shown to asymptotically follow a \textit{Gaussian distribution} with mean and covariance depending on the statistical properties of classes, the ratio of labelled versus unlabelled data, and the value of the first derivatives of the kernel function at the limiting value $\tau$ of $\frac1p\|x_{i}-x_{j}\|^{2}$ (which we recall is irrespective of the genuine classes of $x_i,x_j$). This last finding notably allows one to \textit{predict the asymptotic performances} of the semi-supervised learning algorithms.
\item From the latter result, three main outcomes unfold:
	\begin{itemize}
		\item when three classes or more are considered, there exist Gaussian mixture models for which classification is shown to be \textit{impossible};
		\item despite PageRank's consistency, we further justify that the choice $\alpha=-1$ is not in general optimal. For the case of $2$-class learning, we provide a method to approach the optimal value of $\alpha$; this method is demonstrated on real datasets to convey sometimes \textit{dramatic improvements} in correct classification rates.
		\item for a $2$-class learning task, necessary and sufficient conditions for asymptotic consistency are: $f'(\tau)<0$, $f''(\tau)>0$ and $f''(\tau)f(\tau)>f'(\tau)^2$; in particular, Gaussian kernels, failing to meet the last condition, cannot deal with the large dimensional version of the ``concentric spheres" task.
	\end{itemize}
\end{itemize}

Throughout the article, theoretical results and related discussions are confirmed and illustrated with simulations on Gaussian-mixture data as well as the popular MNIST dataset \citep{lecun1998mnist}, which serves as a comparison for our theoretical study on real world datasets. The consistent match of our theoretical findings on MNIST data, despite their departing from the (very large dimensional) Gaussian mixture assumption, suggests that our results have a certain robustness to this assumption and can be applied to a larger range of datasets.

\medskip

\textit{Notations:} ${\bm\delta}_a^b$ is a binary function taking the value of $1$ if $a=b$ or that of $0$ if not. $1_{n}$ is the column vector of ones of size $n$, $I_{n}$ the $n\times n$ identity matrix. The norm $\|\cdot\|$ is the Euclidean norm for vectors
and the operator norm for matrices. The operator ${\rm diag}(v)={\rm diag}\{v_a\}_{a=1}^k$ is the diagonal matrix having $v_1,\ldots,v_k$ as its ordered diagonal elements. $O(\cdot)$ is the same as specified in \citep{couillet2015kernel}: for a random variable $x\equiv x_n$ and $u_n\ge 0$, we write $x=O(u_n)$ if for any $\eta>0$ and $D>0$, we have $n^D{\rm P}(x\ge n^\eta u_n)\to 0$. When multidimensional objects are concerned, for a vector (or a diagonal matrix)$v$, $v=O(u_n)$ means the maximum entry in absolute value is $O(u_n)$ and for a square matrix $M$, $M=O(u_n)$ means that the operator norm of $M$ is $O(u_n)$. 

\label{sec:approaches}
\section{The optimization framework}
\label{sec:optimization}
Let $x_{1},\ldots,x_{n}\in\mathbb{R}^{p}$ be $n$ data vectors belonging to $K$ classes ${\mathcal C}_{1},\ldots,{\mathcal C}_{K}$. The class association of the $n_{[l]}$ vectors $x_{1},\ldots,x_{n_{[l]}}$ is known (these vectors will be referred to as \textit{labelled}), while the class of the remaining $n_{[u]}$ vectors $x_{n_{[l]}+1},\ldots,x_{n}$ ($n_{[l]}+n_{[u]}=n$) is unknown (these are referred to as \textit{unlabelled} vectors). Within both labelled and unlabelled subsets, the data are organized in such a way that the $n_{[l]1}$ first vectors $x_1,\ldots,x_{[n_{[l]1]}}$ belong to class $\mathcal C_1$, $n_{[l]2}$ subsequent vectors to $\mathcal C_2$, and so on, and similarly for the $n_{[u]1},n_{[u]2},\ldots$ first vectors of the set $x_{n_{[l]}+1},\ldots,x_{n}$. Note already that this ordering is for notational convenience and shall not impact the generality of our results.

The affinity relation between the vectors $x_1,\ldots,x_n$ is measured from the weight matrix $W$ defined by 
\begin{align*}
	W &\equiv \left\{ f\left( \frac1p\|x_i-x_j\|^2 \right) \right\}_{i,j=1}^n
\end{align*}
for some function $f$. The matrix $W$ may be seen as the adjacency matrix of the $n$-node graph indexed by the vectors $x_1,\ldots,x_n$. We further denote by $D$ the diagonal matrix with $D_{ii}\equiv d_{i}=\sum_{j=1}^{n}W_{ij}$ the degree of the node associated to $x_{i}$. 

We next define a score matrix $F\in\mathbb{R}^{n\times K}$ with $F_{ik}$ representing the evaluated score for $x_{i}$ to belong to ${\mathcal C}_{k}$. In particular, following the conventions typically used in graph-based semi-supervised learning \citep{chapelle2006ssl}, we shall affect a unit score $F_{ik}=1$ if $x_i$ is a labelled data of class $\mathcal C_k$ and a null score for all $F_{ik'}$ with $k'\neq k$. In order to attribute classes to the unlabelled data, scores are first affected by means of the resolution of an optimization framework. We propose here
\begin{align}
	\label{eq:opt}
	F &= \argmin_{F\in\RR^{n\times k}} \sum_{k=1}^K \sum_{i,j=1}^n W_{ij} \left\| d_i^\alpha F_{ik} - d_j^\alpha F_{jk} \right\|^2 \nonumber \\
	&\text{s.t.}~ ~ F_{ik}=\begin{cases}
1,& \text{if}\ x_{i}\in\mathcal{C}_{k}\\
0,& \text{otherwise}\
\end{cases},\thinspace1\leq i\leq n_{[l]},\thinspace1\leq k\leq K
\end{align}
where $\alpha\in\RR$ is a given parameter. The interest of this generic formulation is that it coincides with the standard Laplacian-based approach for $\alpha=0$ and with the normalized Laplacian-based approach for $\alpha=-\frac12$, both discussed in Section~\ref{sec:intro}. Note importantly that Equation~\eqref{eq:opt} is naturally motivated by the observation that large values of $W_{ij}$ enforce close values for $F_{ik}$ and $F_{jk}$ while small values for $W_{ij}$ allow for more freedom in the choice of $F_{ik}$ and $F_{jk}$.

By denoting
\begin{align*}
	F &= \begin{bmatrix}F_{[l]} \\F_{[u]}\end{bmatrix},~W=\begin{bmatrix}W_{[ll]} & W_{[lu]}\\W_{[ul]} & W_{[uu]}\end{bmatrix},\text{ and }D=\begin{bmatrix}D_{[l]} & 0\\0 & D_{[u]}\end{bmatrix}
\end{align*}
with $F_{[l]}\in\mathbb{R}^{n_{[l]}}$, $W_{[ll]}\in\mathbb{R}^{n_{[l]}\times n_{[l]}}$, $D_{[l]}\in\mathbb{R}^{n_{[l]}\times n_{[l]}}$, one easily finds (since the problem is a convex quadratic optimization with linear equality constraints) the solution to \eqref{eq:opt} is explicitly given by
\begin{align}
\label{eq:opt-result}
F_{[u]}&=\left(I_{n_{u}}-D_{[u]}^{-1-\alpha}W_{[uu]}D_{[u]}^{\alpha}\right)^{-1}D_{[u]}^{-1-\alpha}W_{[ul]}D_{[l]}^{\alpha}F_{[l]}.
\end{align}
Once these scores are affected, a mere comparison between all scores $F_{i1},\ldots,F_{iK}$ for unlabelled data $x_i$ (i.e., for $i>n_{[l]}$) is performed to decide on its class, i.e., the allocated class index $\hat{\mathcal C}_{x_i}$ for vector $x_i$ is given by
\begin{align*}
	\hat{\mathcal C}_{x_i} &= \mathcal C_{\hat k}\text{ for }\hat k=\argmax_{1\leq k\leq K} F_{ik}.
\end{align*}

\medskip

Note in passing that the formulation \eqref{eq:opt-result} implies in particular that
\begin{align}
	\label{eq:updateFu}
	F_{[u]} &= D_{[u]}^{-1-\alpha}W_{[uu]}D_{[u]}^{\alpha} F_{[u]} + D_{[u]}^{-1-\alpha}W_{[ul]}D_{[l]}^{\alpha}F_{[l]} \\
	\label{eq:updateFl}
	F_{[l]} &= \left\{ {\bm \delta}_{x_i\in\mathcal C_k} \right\}_{\substack{1\leq i\leq n_{[l]} \\ 1\leq k\leq K}}
\end{align}
and thus the matrix $F$ is a stationary point for the algorithm constituted of the updating rules \eqref{eq:updateFu} and \eqref{eq:updateFl} (when replacing the equal signs by affectations). In particular, for $\alpha=-1$, the algorithm corresponds to the standard label propagation method found in the PageRank algorithm for semi-supervised learning as discussed in \citep{avrachenkov2011generalized}, with the major difference that $F_{[l]}$ is systematically reset to its known value where in \citep{avrachenkov2011generalized} $F_{[l]}$ is allowed to evolve (for reasons related to robustness to pre-labeling errors).

The technical objective of the article is to analyze the behavior of $F_{[u]}$ in the large $n,p$ regime for a Gaussian mixture model for the data $x_1,\ldots,x_n$. To this end, we shall first need to design appropriate growth rate conditions for the Gaussian mixture statistics as $p\to\infty$ (in order to avoid trivializing the classification problem as $p$ grows large) before proceeding to the evaluation of the behavior of $W$, $D$, and thus $F$.

\section{Model and Theoretical Results}
\label{sec:model}
\subsection{Model and Assumptions}
In the remainder of the article, we shall assume that the data $x_1,\ldots,x_n$ are extracted from a Gaussian mixture model composed of $K$ classes. Specifically, for $k\in\{1,\ldots,K\}$, 
\begin{align*}
	x_i\in \mathcal C_k \Leftrightarrow x_{i}\sim\mathcal N(\mu_{k},C_{k}).
\end{align*}
Consistently with the previous section, for each $k$, there are $n_{k}$ instances of vectors of class $\mathcal{C}_{k}$, among which $n_{[l]k}$ are labelled and $n_{[u]k}$ are unlabelled.

As pointed out above, in the regime where $n,p\to\infty$, special care must be taken to ensure that the classes $\mathcal C_1,\ldots,\mathcal C_K$, the statistics of which evolve with $p$, remain at a ``somewhat constant'' distance from each other. This is to ensure that the classification problem does not become asymptotically infeasible nor trivially simple as $p\to\infty$. Based on the earlier work \citep{couillet2015kernel} where similar considerations were made, the behavior of the class means, covariances, and cardinalities will follow the prescription below:
\begin{assumption}[\label{Growth-Rate}Growth Rate]
	As $n\to\infty$, $\frac{p}{n}\rightarrow c_{0}>0$ and $\frac{n_{[l]}}n\to c_{[l]}>0$, $\frac{n_{[u]}}n\to c_{[u]}>0$. For each $k$, $\frac{n_{k}}{n}\to c_{k}>0$, $\frac{n_{[l]k}}{n}\to c_{[l]k}>0$, $\frac{n_{[u]k}}{n}\to c_{[u]k}>0$. Besides,
\end{assumption}
\begin{enumerate}
\item For $\mu^{\circ}\triangleq\sum_{k=1}^{K}\frac{n_{k}}{n}\mu_{k}$
and $\mu_{k}^{\circ}\triangleq\mu_{k}-\mu^{\circ}$, $\|\mu_{k}^{\circ}\|=O(1)$.
\item For $C^{\circ}\triangleq\sum_{k=1}^{K}\frac{n_{k}}{n}C_{k}$ and
$C_{k}^{\circ}\triangleq C_{k}-C^{\circ}$, $\|C_{k}\|=O(1)$ and $\text{{\rm tr}}C_{k}^{\circ}=O(\sqrt{p})$.
\item As $n\to\infty$, $\frac{2}{p}{\rm tr}C^{\circ}\rightarrow\tau\neq0$.
\item As $n\to\infty$, $\alpha=O(1)$.
\end{enumerate}
It will also be convenient in the following to define
\begin{align*}
	t_k &\equiv \frac1{\sqrt{p}}\tr C_k^\circ \\
	T_{kk'} &\equiv \frac1p\tr C_kC_{k'}
\end{align*}
as well as the labelled-data centered notations
\begin{align*}
	\tilde \mu_k &\equiv \mu_k - \sum_{k'=1}^K\frac{n_{[l]k'}}{n_{[l]}}\mu_{k'} \\
	\tilde C_k &\equiv C_k - \sum_{k'=1}^K\frac{n_{[l]k'}}{n_{[l]}}C_{k'} \\
	\tilde t_{k} &\equiv \frac1{\sqrt{p}}\tr \tilde C_k \\\
	\tilde T_{kk'} &\equiv \frac1p\tr \tilde C_k\tilde C_{k'}.
\end{align*}

A few comments on Assumption~\ref{Growth-Rate} are in order. First note that, as $n_{[l]}/n$ and $n_{[u]}/n$ remain away from zero, we assume a regime where the number of labelled data is of the same order of magnitude as that of the unlabelled data (although in practice we shall often take $n_{[l]}$ a small, but non vanishing, fraction of $n_{[u]}$). Similarly, the number of representatives of all classes, labelled or not, are of a similar order.

Item 3.\@ is mostly a technical convenience that shall simplify our analysis, ut our results naturally extend as long as both liminf and limsup of $\frac{2}{p}{\rm tr}C^{\circ}$ are away from zero or infinity. The necessity of Item 1.\@ only appears through a detailed analysis of spectral properties of the weight matrix $W$ for large $n,p$, carried out later in the article. As for Item 2.\@, note that if $\tr C_k^\circ=O(\sqrt{p})$ were to be relaxed, it is easily seen that a mere (unsupervised) comparison of the values of $\|x_i\|^2$ would asymptotically provide an almost surely perfect classification.

\medskip

As a by-product of imposing the growth constraints on the data to ensure non-trivial classification, Assumption~\ref{Growth-Rate} induces the following seemingly unsettling implication, easily justified by a simple concentration of measure argument
\begin{align}
	\label{eq:convergence_tau}
	\max_{1\leq i,j\leq n} \left|\frac1p\|x_{i}-x_{j}\|^{2} - \tau\right| \asto 0
\end{align}
as $p\to\infty$. Equation~\eqref{eq:convergence_tau} is the cornerstone of our analysis and states that all vector pairs $x_i,x_j$ are essentially at the same distance from one another as $p$ gets large, \textit{irrespective of their classes}. This striking result evidently is in sharp opposition to the very motivation for the optimization formulation \eqref{eq:opt} as discussed in the introduction. It thus immediately entails that the solution \eqref{eq:opt-result} to \eqref{eq:opt} is bound to produce asymptotically inconsistent results. We shall see that this is indeed the case for all but a short range of values of $\alpha$.

This being said, Equation~\eqref{eq:convergence_tau} has an advantageous side as it allows for a Taylor expansion of $W_{ij}=f(\frac1p\|x_{i}-x_{j}\|^{2})$ around $f(\tau)$, provided $f$ is sufficiently smooth around $\tau$, which is ensured by our subsequent assumption.
\begin{assumption}[\label{Kernel-function}Kernel function]
The function $f\thinspace:\thinspace\mathbb{R}^{+}\rightarrow\mathbb{R}$ is three-times continuously differentiable in a neighborhood of $\tau$.
\end{assumption}
Note that Assumption~\ref{Kernel-function} does not constrain $f$ aside from its local behavior around $\tau$. In particular, we shall not restrict ourselves to matrices $W$ arising from nonnegative definite kernels as standard machine learning theory would advise \citep{scholkopf2002learning}.

The core technical part of the article now consists in expanding $W$, and subsequently all terms intervening in \eqref{eq:opt-result}, in a Taylor expansion of successive matrices of \textit{non-vanishing operator norm}. Note indeed that the magnitude of the individual entries in the Taylor expansion of $W$ needs not follow the magnitude of the operator norm of the resulting matrices;\footnote{For instance, $\|I_n\|=1$ while $\|1_n1_n^\trans\|=n$ despite both matrices having entries of similar magnitude.} rather, great care must be taken to only retain those matrices of non-vanishing operator norm. These technical details call for advanced random matrix considerations and are discussed in the appendix and \citep{couillet2015kernel}.

We are now in position to introduce our main technical results.

\subsection{Main Theoretical Results}
\label{sec:results}

In the course of this section, we provide in parallel a series of technical results under the proposed setting (notably under Assumption~\ref{Growth-Rate}) along with simulation results both on a $2$-class Gaussian mixture data model with $\mu_{1}=[4;0_{p-1}]$, $\mu_{2}=[0;4;0_{p-2}]$, $C_1=I_p$ and $\{C_2\}_{i,j}=.4^{\vert i-j\vert}(1+\frac{3}{\sqrt{p}})$, as well as on real datasets, here images of eights and nines from the MNIST database \citep{lecun1998mnist}, for $f(t)=\exp(-\frac12t)$, i.e., the classical Gaussian (or heat) kernel. For reasons that shall become clear in the following discussion, these figures will depict the (size $n$) vectors
\begin{align*}
	\left[F^\circ_{[u]}\right]_{\cdot k} &\equiv \left[F_{[u]}\right]_{\cdot k} - \frac{1}{K}\sum_{k'=1}^K \left[ F_{[u]} \right]_{\cdot k'}
\end{align*}
for $k\in\{1,2\}$. Obviously, the decision rule on $F^\circ_{[u]}$ is the same as that on $F_{[u]}$.

Our first hinging result concerns the behavior of the score matrix $F$ in the large $n,p$ regime, as per Assumption~\ref{Growth-Rate}, and reads as follows. 
\begin{proposition}
	\label{prop:first-orders-of-Fu} Let Assumptions~\ref{Growth-Rate}--\ref{Kernel-function} hold. Then, for $i>n_{[l]}$ (i.e., for $x_i$ an unlabelled vector),
\begin{equation}\label{eq:first-orders}
	F_{ik}=\frac{n_{[l]k}}n\Bigg[1+\underbrace{(1+\alpha)\frac{\fftau}{\ftau}\frac{t_{k}}{\sqrt{p}}+z_{i}}_{O(n^{-\frac{1}{2}})}+O(n^{-1})\Bigg]
\end{equation}
where $z_{i}=O(n^{-\frac12})$ is a random variable, function of $x_i$, but independent of $k$.
\end{proposition}
The proof of Proposition~\ref{prop:first-orders-of-Fu} is given as an intermediary result of the proof of Theorem~\ref{thm:behaviour of output Fu} in the appendix.

\begin{figure}[h!]
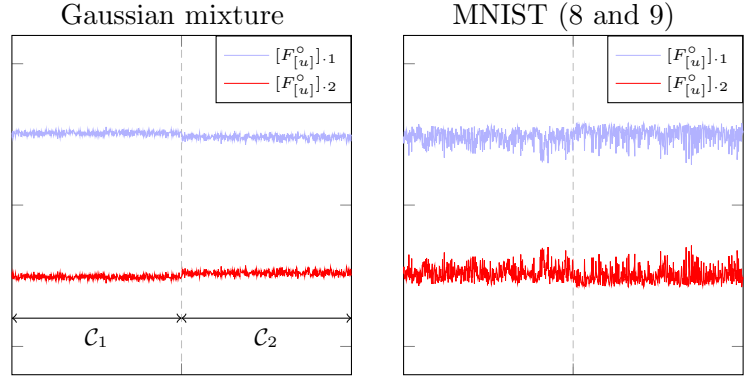

	\centering
\begin{tabular}{cc}
		Gaussian mixture & MNIST (8 and 9) \\
\end{tabular}
	\caption{$[F^\circ_{[u]}]_{\cdot 1}$ and $[F^\circ_{[u]}]_{\cdot 2}$ for $2$-class data, $n=1024$, $p=784$, $n_{l}/n=1/16$, $n_{[u]1}=n_{[u]2}$, $n_{[l]1}=3n_{[l]2}$, $\alpha=-1$, Gaussian kernel.}\label{scores_biased_labelleddata}
\end{figure}

Proposition~\ref{prop:first-orders-of-Fu} provides a clear overview of the outcome of the semi-supervised learning algorithm. First note that $F_{ik}=c_{[l]k}+O(n^{-\frac12})$. Therefore, irrespective of $x_i$, $F_{ik}$ is strongly biased towards $c_{[l]k}$. If the values $n_{[l]1},\ldots,n_{[l]k}$ differ by $O(n)$, this induces a systematic asymptotic allocation of every $x_i$ to the class having largest $c_{[l]k}$ value. Figure~\ref{scores_biased_labelleddata} illustrates this phenomenon, observed both on synthetic and real datasets, here for $n_{[l]1}=3n_{[l]2}$.

\begin{figure}[h!]
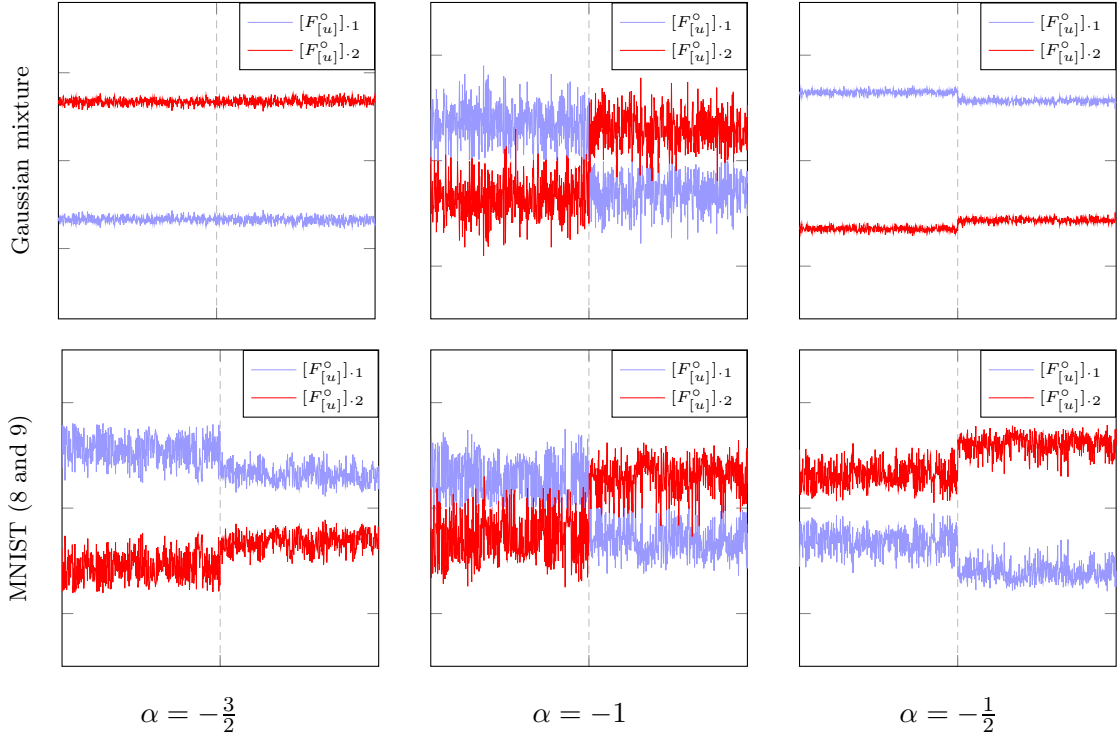

	\centering{}

	\caption{$[F^\circ_{[u]}]_{\cdot 1}$, $[F^\circ_{[u]}]_{\cdot 2}$ for $2$-class data, $n=1024$, $p=784$, $n_{l}/n=1/16$, $n_{[u]1}=n_{[u]2}$, $n_{[l]1}=n_{[l]2}$, Gaussian kernel.}\label{scores_different_alpha}
\end{figure}

Pursuing the analysis of Proposition~\ref{prop:first-orders-of-Fu} by now assuming that $n_{[l]1}=\ldots=n_{[l]K}$, the comparison between $F_{i1},\ldots,F_{iK}$ next revolves around the term of order $O(n^{-\frac12})$. Since $z_i$ only depends on $x_i$ and not on $k$, it induces a constant offset to the vector $F_{i\cdot}$, thereby not intervening in the class allocation. On the opposite, the term $t_k$ is independent of $x_i$ but may vary with $k$, thereby possibly intervening in the class allocation, again an undesired effect. Figure~\ref{scores_different_alpha} depicts the effect of various choices of $\alpha$ for equal values of $n_{[l]k}$. This deleterious outcome can be avoided either by letting $f'(\tau)=O(n^{-\frac12})$ or $\alpha=-1+O(n^{-\frac12})$.  But, as discussed in \citep{couillet2015kernel} and later in the article, the choice of $f$ such that $f'(\tau)\simeq 0$, if sometimes of interest, is generally inappropriate. 

\medskip

The discussion above thus induces two important consequences to adapt the semi-supervised learning algorithm to large datasets.
\begin{enumerate}
	\item The final comparison step \textit{must} be made upon the normalized scores
		\begin{align}
			\label{eq:normalized_scores}
			\hat{F}_{ik} &\equiv \frac{n}{n_{[l]k}} F_{ik}
		\end{align}
		rather than upon the scores $F_{ik}$ directly.
	\item The parameter $\alpha$ \textit{must} be chosen in such a way that
		\begin{align*}
			\alpha &= -1 + O(n^{-\frac12}).
		\end{align*}
\end{enumerate}

\bigskip

Under these two amendments of the algorithm, according to Proposition~\ref{prop:first-orders-of-Fu}, the performance of the semi-supervised learning algorithm now relies upon terms of magnitude $O(n^{-1})$, which are so far left undefined. A thorough analysis of these terms allows for a complete understanding of the asymptotic behavior of the normalized scores $\hat{F}_{i\cdot}=(\hat F_{i1},\ldots,\hat F_{iK})$, as presented in our next result. 
\begin{theorem}
	\label{thm:main}
	Let Assumptions~\ref{Growth-Rate}--\ref{Kernel-function} hold. For $i>n_{[l]}$ (i.e., $x_i$ unlabelled) with $x_i\in\mathcal C_b$, let $\hat F_{ia}$ be given by \eqref{eq:normalized_scores} with $F$ defined in \eqref{eq:opt-result} and $\alpha=-1+\frac{\beta}{\sqrt{p}}$ for $\beta=O(1)$. Then, 
	\begin{align}
		\label{eq:phatF}
	p\hat F_{i\cdot} &= p(1+z_i)1_K+ G_i + o_P(1)
	\end{align}
	where $z_i=O(\sqrt{p})$ is as in Proposition~\ref{prop:first-orders-of-Fu} and $G_i \sim \mathcal N( m_b , \Sigma_b )$, $i>n_{[l]}$, are independent with
	\begin{align}
	[m_b]_a &= -\frac{2f'(\tau)}{f(\tau)} \tilde\mu_a^\trans\tilde\mu_b+\left( \frac{f''(\tau)}{f(\tau)} - \frac{f'(\tau)^2}{f(\tau)^2} \right) \tilde t_a \tilde t_b + \frac{2f''(\tau)}{f(\tau)}\tilde T_{ab} + \frac{\beta}{c_{[l]}} \frac{f'(\tau)}{f(\tau)}t_a \label{eq:thm:main mean}\\
	[\Sigma_b]_{a_1a_2} &= 2\left( \frac{f''(\tau)}{f(\tau)} - \frac{f'(\tau)^2}{f(\tau)^2} \right)^2T_{bb} t_{a_1}t_{a_2} +4\frac{f'(\tau)^2}{f(\tau)^2} \left[ \mu_{a_1}^{\circ\trans}C_b\mu_{a_2}^\circ + {\bm\delta}_{a_1}{a_2} \frac{c_0 T_{b,a_1}}{c_{[l]}c_{[l]a_1}} \right]. \label{eq:thm:main covariance}
	\end{align}
	Besides, there exists $\mathcal A\subset \sigma(\{\{x_1,\ldots,x_{n_{[l]}}\},p=1,2,\ldots\})$ (the $\sigma$-field induced by the labelled variables) with $P(\mathcal A)=1$ over which \eqref{eq:phatF} also holds conditionally to $\{\{x_1,\ldots,x_{n_{[l]}}\},p=1,2,\ldots\}$.
\end{theorem}
Note that the statistics of $G_i$ are independent of the realization of $x_1,\ldots,x_{[l]}$ when $\alpha=-1+O(\frac{1}{\sqrt{p}})$. This in fact no longer holds when $\alpha$ is outside this regime, as pointed out by Theorem~\ref{thm:behaviour of output Fu} in the appendix which provides the asymptotic behavior of $\hat{F}_{i\cdot}$ for all values of $\alpha$ (and thus generalizes Theorem~\ref{thm:main}).

Since the ordering of the entries of $\hat F_{i\cdot}$ is the same as that of $\hat F_{i\cdot}-(1+z_i)$,  Theorem~\ref{thm:main} amounts to saying that the probability of correctly classifying unlabeled vectors $x_i$ genuinely belonging to class $\mathcal C_b$ is asymptotically given by the probability of $[G_{i}]_b$ being the maximal element of $G_i$, which, as mentioned above,  is the same whether conditioned or not on $x_1,\ldots,x_{[l]}$ for $\alpha=-1+O(\frac{1}{\sqrt{p}})$. This is formulated in the following corollary.
\begin{corollary}
	\label{cor:proba}
	Let Assumptions~\ref{Growth-Rate}--\ref{Kernel-function} hold. Let $i>n_{[l]}$ and $\alpha=-1+\frac{\beta}{\sqrt{p}}$. Then, under the notations of Theorem~\ref{thm:main}, 
	\begin{align*}
	{\mathbb P}\left(x_{i}\to\mathcal{C}_{b}|x_{i}\in\mathcal{C}_{b},x_{1},\cdots,x_{n_{[l]}}\right) - {\mathbb P}\left(x_{i}\to\mathcal{C}_{b}|x_{i}\in\mathcal{C}_{b}\right)  &\to 0\\
	{\mathbb P}\left(x_{i}\to\mathcal{C}_{b}|x_{i}\in\mathcal{C}_{b}\right) - {\mathbb P}\left([G_{i}]_b>\max_{a\neq b} \{[G_i]_a\}|x_{i}\in\mathcal{C}_{b}\right) &\to0.
	\end{align*}
	In particular, for $K=2$, and $a\neq b\in\{1,2\}$,
	\begin{align*}
	&{\mathbb P}\left([G_{i}]_b>\max_{a\neq b} \{[G_i]_a\}|x_{i}\in\mathcal{C}_{b}\right)= \Phi(\theta_b^a),\quad \textmd{with }\theta_b^a \equiv \frac{[m_b]_b-[m_b]_a}{\sqrt{[\Sigma_b]_{bb}+[\Sigma_b]_{aa}-2[\Sigma_b]_{ab}}}
	\end{align*}
	where $\Phi(u)=\frac1{2\pi}\int_{-\infty}^ue^{-\frac{t^2}2}dt$ is the Gaussian distribution function.
\end{corollary}

With $G_i$ being independent, Corollary~\ref{cor:proba} allows us to approach the empirical classification accuracy as it is consistently estimated by the probability of correct classification given in the corollary. As with Theorem~\ref{thm:main} which can be appended to Theorem~\ref{thm:behaviour of output Fu} for a large set of values of $\alpha$, Corollary~\ref{cor:proba} is similarly generalized by Corollary~\ref{cor:conditional precision} in the appendix. Using both corollaries, Figure~\ref{fig:acc_MINST} displays a comparison between simulated accuracies from various pairs of digits from the MNIST dataset against our theoretical results; to apply our results, a $2$-class Gaussian mixture model is assumed with means and covariances equal to the empirical means and covariances of the individual digits, evaluated from the full $60\,000$-image MNIST database. 
It is quite interesting to observe that, despite the obvious inadequacy of a Gaussian mixture model for this image dataset, the theoretical predictions are in strong agreement with the practical performances. Also surprising is the strong adequacy of the theoretical prediction of Corollary~\ref{cor:proba} beyond the range of values of $\alpha$ in the neighborhood of $-1$.
\begin{figure}[h!]
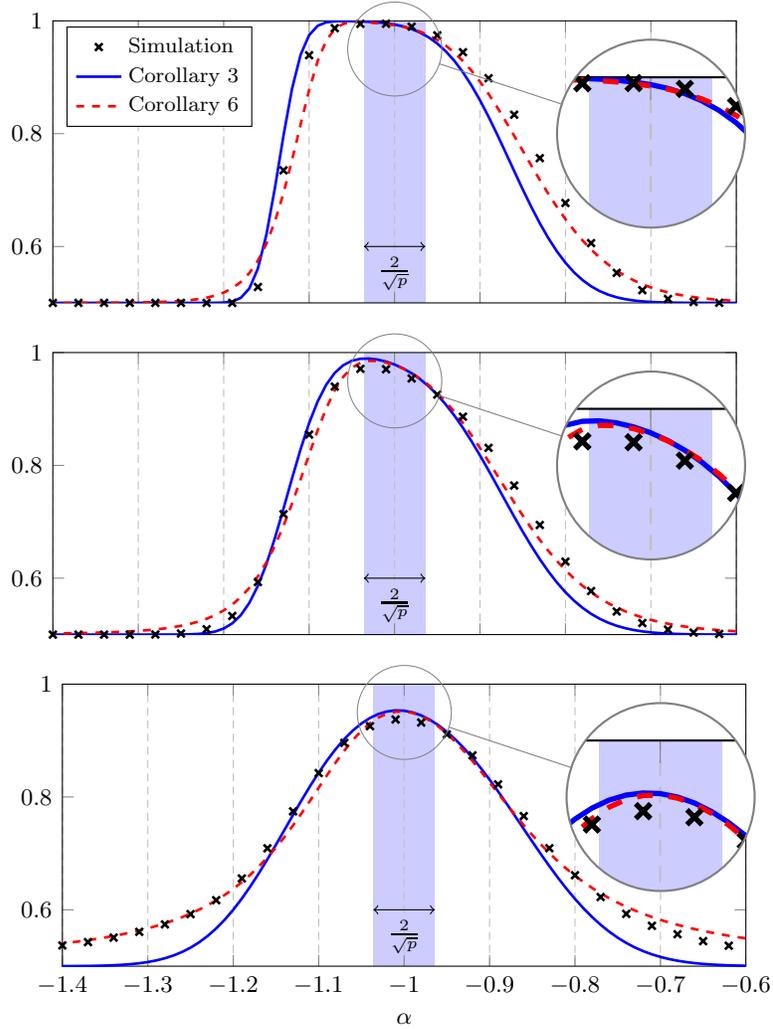

	\centering

	\caption{Theoretical and empirical accuracy as a function of $\alpha$ for 2-class MNIST data ({\bf top:} digits (0,1), {\bf middle:} digits (1,7), {\bf bottom:} digits (8,9)), $n=1024$, $p=784$, $n_{[l]}/n=1/16$, $n_{[u]1}=n_{[u]2}$, Gaussian kernel. Averaged over $50$ iterations.} 
	\label{fig:acc_MINST}
\end{figure}

\section{Consequences}

\subsection{Semi-supervised learning beyond two classes}

An immediate consequence of Corollary~\ref{cor:proba} is that, for $K>2$, there exists a Gaussian mixture model for which the semi-supervised learning algorithms under study necessarily fail to classify at least one class. To see this, it suffices to consider $K=3$ and let $\mu_3=3\mu_2=6\mu_1$, $C_1=C_2=C_3$, $n_1=n_2=n_3$, $n_{[l]1}=n_{[l]2}=n_{[l]3}$. First, it follows from Corollary~\ref{cor:proba} that,
\begin{align*}
	&{\mathbb P}\left(x_{i}\to\mathcal{C}_{2}|x_{i}\in\mathcal{C}_{2}\right)\leq{\mathbb P}\left([G_{i}]_2>[G_i]_1|x_{i}\in\mathcal{C}_{2}\right)+o(1)=\Phi(\theta_2^1)+o(1)\\
	&{\mathbb P}\left(x_{i}\to\mathcal{C}_{3}|x_{i}\in\mathcal{C}_{3}\right)\leq{\mathbb P}\left([G_{i}]_3>[G_i]_1|x_{i}\in\mathcal{C}_{3}\right)+o(1)=\Phi(\theta_3^1)+o(1)
\end{align*}
Then, under Assumptions~\ref{Growth-Rate}--\ref{Kernel-function} and the notations of Corollary~\ref{cor:proba},
\begin{align*}
	\theta_2^1 &= {\rm sgn}(f'(\tau)) \frac{\mu_1^2}{\sqrt{(\Sigma_2)_{22}+(\Sigma_2)_{11}-2(\Sigma_2)_{12}}} \\
	\theta_3^1 &= -{\rm sgn}(f'(\tau)) \frac{15\mu_1^2}{\sqrt{(\Sigma_3)_{33}+(\Sigma_3)_{11}-2(\Sigma_3)_{13}}}
\end{align*}
so that $f'(\tau)<0\Rightarrow \theta_2^1<0$, $f'(\tau)>0\Rightarrow \theta_3^1<0$, while $f'(\tau)=0\Rightarrow \theta_2^1=\theta_3^1=0$. As such, the correct classification rate of elements of $\mathcal C_2$ and $\mathcal C_3$ cannot be simultaneously greater than $\frac12$, leading to necessarily inconsistent classifications.

\subsection{Choice of $f$ and sub-optimality of the heat kernel}
\label{sec:f}

As a consequence of the previous section, we shall from here on concentrate on the semi-supervised classification of $K=2$ classes. In this case, it is easily seen that,
\begin{align*}
	(K=2)\quad \forall a\neq b\in\{1,2\},~ \|\tilde \mu_b\|^2\geq \tilde \mu_b^\trans \tilde \mu_a,\quad \tilde{t}_b^2 \geq \tilde{t}_a\tilde{t}_b,\quad \tilde T_{bb}\geq \tilde T_{ab}
\end{align*}
with equalities respectively for $\mu_a=\mu_b$, $t_a=t_b$, and $\tr C_aC_b=\tr C_b^2$. This result, along with Corollary~\ref{cor:proba}, implies the necessity of the conditions
\begin{align*}
	f'(\tau)<0, \quad f''(\tau)f(\tau)>f'(\tau)^2,\quad f''(\tau)>0
\end{align*}
to fully discriminate Gaussian mixtures. As such, from Corollary~\ref{cor:proba}, by letting $\alpha=-1$, semi-supervised classification of $K=2$ classes is always consistent under these conditions.

\medskip

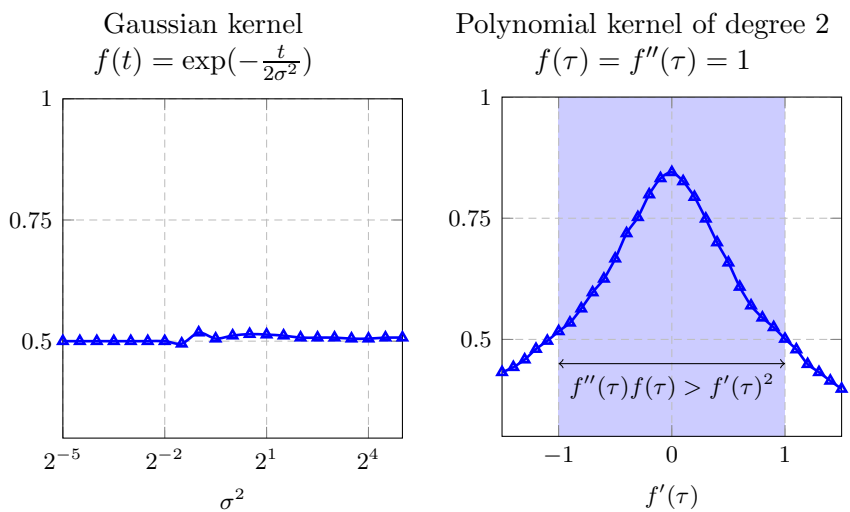
\begin{figure}[h!]
	\centering
	\begin{tabular}{cc}
		Gaussian kernel & Polynomial kernel of degree 2\\
		$f(t)=\exp(-\frac{t}{2\sigma^2})$&$\ftau=\ffftau=1$\\
		\begin{tikzpicture}[font=\footnotesize]
		\renewcommand{\axisdefaulttryminticks}{4} 
		\tikzstyle{every major grid}+=[style=densely dashed]       
		\tikzstyle{every axis y label}+=[yshift=-10pt] 
		\tikzstyle{every axis x label}+=[yshift=5pt]
		\tikzstyle{every axis legend}+=[cells={anchor=west},fill=white,
		at={(1,1)}, anchor=north east, font=\tiny ]
		\begin{axis}[
		width=.4\linewidth,
		height=.4\linewidth,
		ytick={.25,.5,.75,1},
		ymin=0.3,
		ymax=1,
       xmode=log,
       log basis x={2},
       xmin=0.03125,
       xmax=32,
       xminorticks=true,
       xlabel={$\sigma^2$},
		grid=major,
		]
		\addplot[color=blue,smooth,solid,line width=1.0pt,mark size=2.0pt,mark=triangle,mark options={solid}]  coordinates{
	
	(0.031250,0.500000)(0.044194,0.500000)(0.062500,0.500000)(0.088388,0.500000)(0.125000,0.500000)(0.176777,0.500000)(0.250000,0.500000)(0.353553,0.494792)(0.500000,0.517708)(0.707107,0.505208)(1.000000,0.511458)(1.414214,0.514583)(2.000000,0.513542)(2.828427,0.511458)(4.000000,0.507292)(5.656854,0.507292)(8.000000,0.507292)(11.313708,0.505208)(16.000000,0.505208)(22.627417,0.507292)(32.000000,0.507292)
		};
	
		\end{axis}
		\end{tikzpicture} &
		\begin{tikzpicture}[font=\footnotesize]
		\renewcommand{\axisdefaulttryminticks}{4} 
		\tikzstyle{every major grid}+=[style=densely dashed]       
		\tikzstyle{every axis y label}+=[yshift=-10pt] 
		\tikzstyle{every axis x label}+=[yshift=5pt]
		\tikzstyle{every axis legend}+=[cells={anchor=west},fill=white,
		at={(1,1)}, anchor=north east, font=\tiny ]
		\begin{axis}[
		width=.4\linewidth,
		height=.4\linewidth,
		xmin=-1.5,
		ymin=0.3,
		xmax=1.5,
		ymax=1,
		ytick={.25,.5,.75,1},
		xlabel={$\fftau$},
		bar width=1.5pt,
		grid=major,
		scaled ticks=true,
		]
		\addplot[color=blue,smooth,solid,line width=1.0pt,mark size=2.0pt,mark=triangle,mark options={solid}] coordinates{
(-1.500000,0.432292)(-1.400000,0.442708)(-1.300000,0.458333)(-1.200000,0.480208)(-1.100000,0.496875)(-1.000000,0.516667)(-0.900000,0.534375)(-0.800000,0.563542)(-0.700000,0.596875)(-0.600000,0.625000)(-0.500000,0.666667)(-0.400000,0.718750)(-0.300000,0.752083)(-0.200000,0.798958)(-0.100000,0.832292)(0.000000,0.844792)(0.100000,0.826042)(0.200000,0.793750)(0.300000,0.748958)(0.400000,0.700000)(0.500000,0.658333)(0.600000,0.608333)(0.700000,0.569792)(0.800000,0.544792)(0.900000,0.525000)(1.000000,0.501042)(1.100000,0.479167)(1.200000,0.448958)(1.300000,0.432292)(1.400000,0.414583)(1.500000,0.397917)
		};
		\begin{pgfonlayer}{background}
			\fill [color=blue!20] (axis cs:-1,.3) rectangle (axis cs:1,1);
		\end{pgfonlayer}
		\draw[<->] (axis cs:-1,.45) -- (axis cs:1,.45) node [below,pos=.5] {$f''(\tau)f(\tau)>f'(\tau)^2$};

		\end{axis}
		\end{tikzpicture}
	\end{tabular}
	\caption{Empirical accuracy for $2$-class Gaussian data with $\mu_{1}=\mu_{2}$, $C_1=I_p$ and $C_2=(1+\frac{3}{\sqrt{p}})I_p$, $n=1024$, $p=784$, $n_{l}/n=1/16$, $n_{[u]1}=n_{[u]2}$, $n_{[l]1}=n_{[l]2}$, $\alpha=-1$.}\label{fig:classification var}
\end{figure}

Consequently, a quite surprising outcome of the discussion above is that the widely used Gaussian (or heat) kernel $f(t)=\exp(-\frac{t}{2\sigma^2})$, while fulfilling the condition $f'(t)<0$ and $f''(t)>0$ for all $t$ (and thus $f'(\tau)<0$ and $f''(\tau)>0$), only satisfies $f''(t)f(t)=f'(t)^2$. This indicates that discrimination over $t_1,\ldots,t_K$, under the conditions of Assumption~\ref{Growth-Rate}, is asymptotically \textit{not} possible with a Gaussian kernel.
This remark is illustrated in Figure~\ref{fig:classification var} for a discriminative task between two centered isotropic Gaussian classes only differing by the trace of their covariance matrices. There, irrespective of the choice of the bandwidth $\sigma$, the Gaussian kernel leads to a constant $1/2$ accuracy, where a mere second order polynomial kernel selected upon its derivatives at $\tau$ demonstrates good performances. Since $p$-dimensional isotropic Gaussian vectors tend to concentrate ``close to'' the surface of a sphere, this thus suggests that Gaussian kernels are not inappropriate to solve the large dimensional generalization of the ``concentric spheres'' task (for which they are very efficient in small dimensions).
In passing, the right-hand side of Figure~\ref{fig:classification var} confirms the need for $f''(\tau)f(\tau)-f'(\tau)^2$ to be positive (there $|f'(\tau)|<1$) as an accuracy lower than $1/2$ is obtained for $f''(\tau)f(\tau)-f'(\tau)^2<0$. 


\medskip

Another interesting fact lies in the choice $f'(\tau)=0$ (while $f''(\tau)\neq 0$). As already identified in \citep{couillet2015kernel} and further thoroughly investigated in \citep{kammoun2017}, if $t_1=t_2$ (which can be enforced by normalizing the dataset) and $\tilde T_{bb}>\tilde T_{ba}$ for all $a\neq b\in\{1,2\}$, then $\Sigma_b=0$ while $[m_b]_b>[m_b]_a$ for all $b\neq a\in\{1,2\}$ and thus leading asymptotically to a perfect classification.
As such, while Assumption~\ref{Growth-Rate} was claimed to ensure a ``non-trivial'' growth rate regime, asymptotically perfect classification may be achieved by choosing $f$ such that $f'(\tau)=0$, under the aforementioned statistical conditions. One must nonetheless be careful that this \textit{asymptotic result} does not necessarily entail outstanding performances in practical finite dimensional scenarios. Indeed, note that taking $f'(\tau)=0$ discards the visibility of differing means $\mu_1\neq \mu_2$ (from the expression of $[m_b]_a$ in Theorem~\ref{thm:main}); for finite $n,p$, cancelling the differences in means (often larger than differences in covariances) may not be compensated for by the reduction in variance. Trials on MNIST particularly emphasize this remark.

\subsection{Impact of class sizes}

A final remark concerns the impact of $c_{[l]}$ and $c_0$ on the asymptotic performances. Note that $c_{[l]}$ and $c_0$ only act upon the covariance $\Sigma_b$ and precisely on its diagonal elements. Both a reduction in $c_0$ (by increasing $n$) and an increase in $c_{[l]}$ reduce the diagonal terms in the variance, thereby mechanically increasing the classification performances (if in addition $[m_b]_b>[m_b]_a$ for $a\neq b$). This is a naturally expected result, however in general not leading to a vanishing classification error.

\section{Parameter Optimization in Practice}

\subsection{Estimation of $\tau$}

In previous sections, we have emphasized the importance of selecting the kernel function $f$ so as to meet specific conditions on its derivatives at the quantity $\tau$. In practice however, $\tau$ is an unknown quantity. A mere concentration of measure argument nonetheless shows that 
\begin{align}
	\label{eq:hattau}
	\hat\tau \equiv \frac1{n(n-1)}\sum_{i,j=1}^n \frac1p\|x_i-x_j\|^2 \asto \tau.
\end{align}
As a consequence, the results of Theorem~\ref{thm:main} and subsequently of Corollary~\ref{cor:proba} hold verbatim with $\hat\tau$ in place of $\tau$. One thus only needs to design $f$ in such as way that its derivatives at $\hat\tau$ meet the appropriate conditions.


\subsection{Optimization of $\alpha$}

In Section~\ref{sec:f}, we have shown that the choice $\alpha=-1$, along with an appropriate choice of $f$, ensures the asymptotic consistency of semi-supervised learning for $K=2$ classes, in the sense that non-trivial asymptotic accuracies ($>0.5$) can be achieved. This choice of $\alpha$ may however not be optimal in general. This subsection is devoted to the optimization of $\alpha$ so as to maximize the average precision, a criterion often used in absence of prior information to favor one class over the other. While not fully able to estimate the optimal $\alpha^\star$ of $\alpha$, we shall discuss here a heuristic means to select a close-to-optimal $\alpha$, subsequently denoted $\alpha_0$.

\medskip

As per Theorem~\ref{thm:main}, $\alpha$ must be chosen as $\alpha=-1+\frac{\beta}{\sqrt{p}}$ for some $\beta=O(1)$. In order to set $\beta=\beta^\star$ in such a way that the classification accuracy is maximized, Corollary~\ref{cor:proba} further suggests the need to estimate the $\theta_b^a$ terms which in turn requires the evaluation of a certain number of quantities appearing in the expressions of $m_b$ and $\Sigma_b$. Most of these are however not directly accessible from simple statistics of the data. Instead, we shall propose here a heuristic and simple method to retrieve a reasonable choice $\beta_0$ for $\beta$, which we claim is often close to optimal and sufficient for most needs.

To this end, first observe from \eqref{eq:thm:main mean} that the mappings $\beta\mapsto [m_b]_a$ satisfy
\begin{align*}
	\frac{d}{d\beta}([m_b]_b-[m_b]_a) &= \frac{f'(\tau)}{f(\tau)c_{[l]}}(t_b-t_a) = -\frac{d}{d\beta}([m_a]_a-[m_a]_b).
\end{align*}
Hence, changes in $\beta$ induce a simultaneous reduction and increase of either one of $[m_b]_b-[m_b]_a$ and $[m_a]_a-[m_a]_b$. Placing ourselves again in the case $K=2$, we define $\beta_0$ to be the value for which both differences (with $a\neq b\in\{1,2\}$) are the same, leading to the following Proposition--Definition.

\begin{proposition}
	\label{prop:beta0}
	Let $K=2$ and $[m_b]_a$ be given by \eqref{eq:thm:main mean}. Then
	\begin{equation}
		\label{eq:beta opt}
		\beta_0 \equiv \frac{\ftau}{\ffftau} \frac{c_{[l]1}-c_{[l]2}}{t_1-t_2}\Delta m
	\end{equation}	
	where \begin{align*}
		&\Delta m=-\frac{2f'(\tau)}{f(\tau)} \|\mu_1-\mu_2\|^2+\left( \frac{f''(\tau)}{f(\tau)} - \frac{f'(\tau)^2}{f(\tau)^2} \right)(t_1-t_2)^2 + \frac{2f''(\tau)}{f(\tau)}(T_{11}+T_{22}-2T_{12})
	\end{align*}
	is such that, for $\alpha=-1+\frac{\beta_0}{\sqrt{p}}$, $[m_1]_1-[m_1]_2=[m_2]_2-[m_2]_1$.
\end{proposition}

By choosing $\alpha=\alpha_0\equiv -1+\frac{\beta_0}{\sqrt{p}}$, one ensures that $\mathbb{E}_{x_i\in\mathcal C_1}[\hat F_{i1}-\hat F_{i2}]=-\mathbb{E}_{x_i\in\mathcal C_2}[\hat F_{i1}-\hat F_{i2}]+o(1)$ ($i>n_{[l]}$), thereby evenly balancing the {\it average} ``resolution'' of each class. An even balance typically produces the desirable output of the central displays of Figure~\ref{scores_different_alpha} (as opposed to the largely undesirable bottom of top displays, there for very offset values of $\alpha$). Obviously though, since the variances of $\hat F_{i1}-\hat F_{i2}$ for $x_i\in\mathcal C_1$ or $x_i\in\mathcal C_2$ are in general not the same, this choice of $\alpha$ may not be optimal. Nonetheless, in most experimental scenarios of practical interest, the score variances tend to be sufficiently similar for the choice of $\alpha_0$ to be quite appealing.

This heuristic motivation made, note that $\beta_0$ is proportional to $c_{[l]b}-c_{[l]a}$. This indicates that the more unbalanced the labelled dataset the more deviated from zero $\beta_0$. In particular, for $n_{[l]1}=n_{[l]2}$, $\alpha_0=-1$. As we shall subsequently observe in simulations, this remark is of dramatic importance in practice where taking $\alpha=-1$ (the PageRank method) in place of $\alpha=\alpha_0$ leads to significant performance losses.

\medskip

Of utmost importance here is the fact that, unlike $\theta_b^a$ which are difficult to assess empirically, a consistent estimate of $\beta_0$ can be obtained through a rather simple method, which we presently elaborate on.

While an estimate for $t_a$ and $T_{ab}$ can be obtained empirically from the labelled data themselves, $\|\mu_1-\mu_2\|^2$ is not directly accessible (note indeed that $\frac1{n_{[l]a}}\sum_{\mathcal C_a}x_i=\mu_a+\frac1{n_{[l]a}}\sum_{\mathcal C_a}w_i$, for some $w_i\sim\mathcal N(0,C_a)$, and the central limit theorem guarantees that $\|\frac1{n_{[l]a}}\sum_{\mathcal C_a}w_i\|=O(1)$, the same order of magnitude as $\|\mu_a-\mu_b\|$). However, one may access an estimate for $\Delta m$ by running two instances of the PageRank algorithm ($\alpha=-1$), resulting in the method described in Algorithm~\ref{alg:beta0}. It is easily shown that, under Assumptions~\ref{Growth-Rate}--\ref{Kernel-function},
\begin{align*}
	\hat \beta_0 - \beta_0 \asto 0.
\end{align*}

\begin{algorithm}
	\caption{Estimate $\hat{\beta}_0$ of $\beta_0$.}
\label{alg:beta0}
 \begin{algorithmic}[1]
	 \STATE Let $\hat \tau$ be given by \eqref{eq:hattau}.
	 \STATE Let
 \begin{align*}
	 \widehat{\Delta t} = \frac1{2\sqrt{p}} \left( \frac{\sum_{i,j=1}^{n_{[l]1}}\|x_i-x_j\|^2}{n_{[l]1}(n_{[l]1}-1)} - \frac{\sum_{i,j=n_{[l]1}+1}^{n_{[l]}}\|x_i-x_j\|^2}{n_{[l]2}(n_{[l]2}-1)} \right)
 \end{align*}
 \STATE Set $\alpha=-1$ and define $J \equiv p \sum_{i=n_{[l]}+1}^n \hat F_{i1}-\hat F_{i2}$.
 \STATE Still for $\alpha=-1$, reduce the set of labelled data to $n'_{[l]1}=n'_{[l]2}=\min\{n_{[l]1},n_{[l]2}\}$ and, with obvious notations, let $J' \equiv p \sum_{i=n'_{[l]}+1}^n \hat F'_{i1}-\hat F'_{i2}$.
 \STATE Return $\hat\beta_0 \equiv \frac{c_{[l]}f(\hat \tau)}{f'(\hat \tau) \widehat{\Delta t}} \frac{J'-J}{n_{[u]}}$.
\end{algorithmic}
\end{algorithm}

\medskip

\begin{figure}[h!]
	\centering
	\begin{tabular}{c}
		\begin{tikzpicture}[font=\footnotesize]
		\renewcommand{\axisdefaulttryminticks}{4} 
		\tikzstyle{every major grid}+=[style=densely dashed]       
		\tikzstyle{every axis y label}+=[yshift=-10pt] 
		\tikzstyle{every axis x label}+=[yshift=5pt]
		\tikzstyle{every axis legend}+=[cells={anchor=west},fill=white,
		at={(0.98,0.02)}, anchor=south east, font=\scriptsize ]
		\begin{axis}[
		width=.7\linewidth,
		height=.35\linewidth,
		ymin=0.9,
	ymax=1,
	xmode=log,
	log basis x={2},
	xmin=0.0625,
	xmax=0.5,
	xminorticks=true,
	xticklabels={},
	grid=major,]
		\addplot[color=blue,smooth,dashed,line width=1.0pt,mark size=2.0pt,mark=triangle,mark options={solid}] coordinates{
(0.500000,0.996094)(0.353553,0.995313)(0.250000,0.995833)(0.176777,0.997656)(0.125000,0.995833)(0.088388,0.997135)(0.062500,0.996615)};
		\addplot[color=red,smooth,solid,line width=1.0pt,mark size=2.0pt,mark=triangle,mark options={solid}] coordinates{
(0.500000,0.995052)(0.353553,0.990885)(0.250000,0.993229)(0.176777,0.989583)(0.125000,0.992969)(0.088388,0.996094)(0.062500,0.996615)
		};
		\addplot[color=black,smooth,solid,line width=1.0pt,mark size=2.0pt,mark=o,mark options={solid}] coordinates{
(0.500000,0.995052)(0.353553,0.993750)(0.250000,0.990885)(0.176777,0.983594)(0.125000,0.961719)(0.088388,0.925000)(0.062500,0.893750)
		};
		\legend{ {$\alpha=\alpha^\star$ (oracle)},{$\alpha=-1+\frac{\hat{\beta}_0}{\sqrt{p}}$ (Algorithm~\ref{alg:beta0})},{$\alpha=-1$ (PageRank method)}}
		\end{axis}
		\end{tikzpicture}
		\\
		\definecolor{mycolor1}{rgb}{0.00000,1.00000,1.00000}%
		\definecolor{mycolor2}{rgb}{1.00000,0.00000,1.00000}%
		\begin{tikzpicture}[font=\footnotesize]
		\renewcommand{\axisdefaulttryminticks}{4} 
		\tikzstyle{every major grid}+=[style=densely dashed]       
		\tikzstyle{every axis y label}+=[yshift=-10pt] 
		\tikzstyle{every axis x label}+=[yshift=5pt]
		\tikzstyle{every axis legend}+=[cells={anchor=west},fill=white,
		at={(0.98,0.98)}, anchor=north east, font=\scriptsize ]
		\begin{axis}[
		width=.7\linewidth,
		height=.35\linewidth,
		ymin=0.8,
        ymax=1,
        xmode=log,
        log basis x={2},
        xmin=0.0625,
        xmax=0.5,
        xminorticks=true,
	xticklabels={},
        grid=major,
		]
		\addplot[color=blue,smooth,dashed,line width=1.0pt,mark size=2.0pt,mark=triangle,mark options={solid}] coordinates{
(0.500000,0.978646)(0.353553,0.972135)(0.250000,0.977083)(0.176777,0.974219)(0.125000,0.977083)(0.088388,0.978385)(0.062500,0.976302)
};
        \addplot[color=red,smooth,solid,line width=1.0pt,mark size=2.0pt,mark=triangle,mark options={solid}] coordinates{
(0.500000,0.971094)(0.353553,0.971354)(0.250000,0.975000)(0.176777,0.974219)(0.125000,0.973958)(0.088388,0.976302)(0.062500,0.976042)
};
        \addplot[color=black,smooth,solid,line width=1.0pt,mark size=2.0pt,mark=o,mark options={solid}] coordinates{
(0.500000,0.971094)(0.353553,0.938281)(0.250000,0.903646)(0.176777,0.889323)(0.125000,0.857031)(0.088388,0.843490)(0.062500,0.823958)
};
		\end{axis}
		\end{tikzpicture} 
		\\
		\hspace*{.01\linewidth}
		\begin{tikzpicture}[font=\footnotesize]
	\renewcommand{\axisdefaulttryminticks}{4} 
	\tikzstyle{every major grid}+=[style=densely dashed]       
	\tikzstyle{every axis y label}+=[yshift=-10pt] 
	\tikzstyle{every axis x label}+=[yshift=5pt]
	\tikzstyle{every axis legend}+=[cells={anchor=west},fill=white,
	at={(0.98,0.98)}, anchor=north east, font=\scriptsize ]
	\begin{axis}[
	width=.7\linewidth,
	height=.35\linewidth,
	ymin=0.8,
	ymax=1,
	xmode=log,
	log basis x={2},
	xmin=0.0625,
	xmax=0.5,
	xminorticks=true,
	grid=major,
	xlabel={$c_{[l]1}$},
	]
	\addplot[color=blue,smooth,dashed,line width=1.0pt,mark size=2.0pt,mark=triangle,mark options={solid}] coordinates{
(0.500000,0.945573)(0.353553,0.947396)(0.250000,0.953646)(0.176777,0.948177)(0.125000,0.940365)(0.088388,0.934635)(0.062500,0.949479)};
	\addplot[color=red,smooth,solid,line width=1.0pt,mark size=2.0pt,mark=triangle,mark options={solid}] coordinates{
(0.500000,0.945312)(0.353553,0.947396)(0.250000,0.952083)(0.176777,0.948177)(0.125000,0.940365)(0.088388,0.930990)(0.062500,0.942448)
	};
	\addplot[color=black,smooth,solid,line width=1.0pt,mark size=2.0pt,mark=o,mark options={solid}] coordinates{
(0.500000,0.945312)(0.353553,0.935417)(0.250000,0.907552)(0.176777,0.889323)(0.125000,0.849740)(0.088388,0.810417)(0.062500,0.793750)
	};
	\end{axis}
	\end{tikzpicture} 
	\end{tabular}
	\caption{Average precision varying with $c_{[l]1}$ for 2-class MNIST data ({\bf top:} digits (0,1), {\bf middle:} digits (1,7), {\bf bottom:} digits (8,9)), $n=4096$, $p=784$, $n_{[l]}/n=1/16$, $n_{[u]1}=n_{[u]2}$, Gaussian kernel.}
	\label{fig:beta_MNIST}
\end{figure}
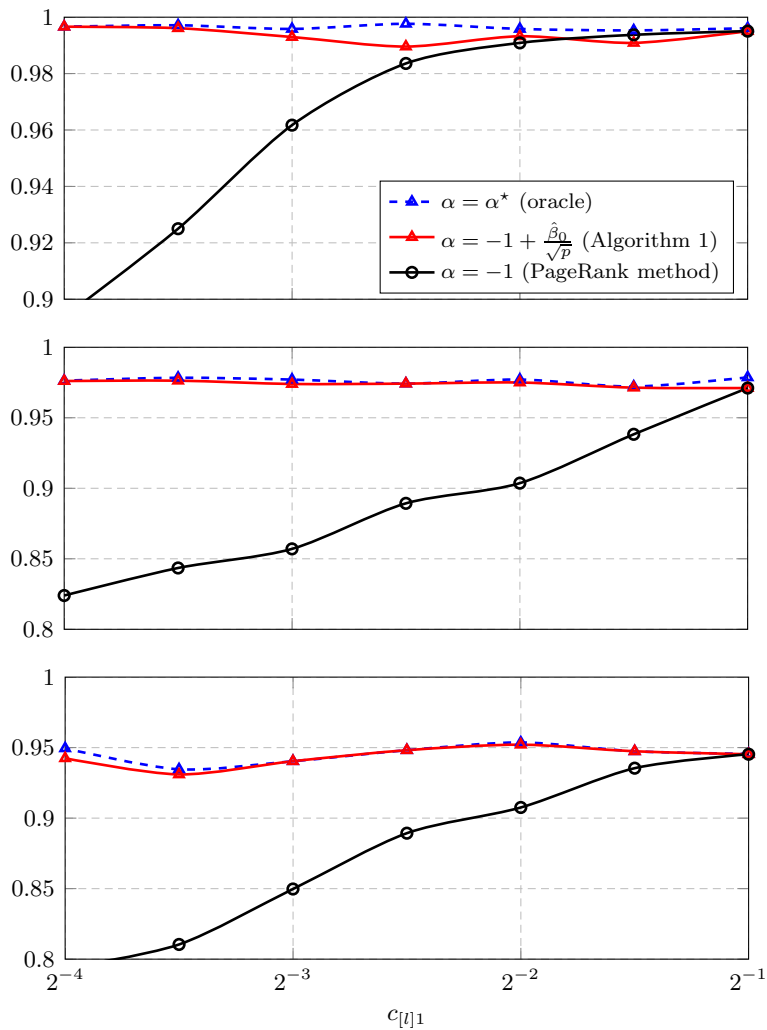

Figure~\ref{fig:beta_MNIST} provides a performance comparison, in terms of average precision, between the PageRank ($\alpha=-1$) method and the proposed heuristic improvement for $\alpha=\alpha_0$, versus the oracle estimator for which $\alpha=\alpha^\star$, the precision-maximizing value. The curves are here snapshots of typical classification precisions obtained from examples of $n=4096$ images with $c_{[l]}=1/16$. As expected, the gain in performance is largest as $|c_{[l]1}-c_{[l]2}|$ is large. More surprisingly, the performances obtained are impressively close to optimal. It should be noted though that simulations revealed more unstable estimates of $\hat\beta_0$ for smaller values of $n$.

\medskip

Note that the method for estimating $\beta_0$ provided in Algorithm~\ref{alg:beta0} implicitly exploits the resolution of two equations (through the observation of $J,J'$ obtained for different values of $n_{[l]1},n_{[l]2}$) to retrieve the value of $\Delta m$ defined in Proposition~\ref{prop:beta0}. Having access to $\Delta m$ further allows access to $\|\mu_1-\mu_2\|^2$, for instance by setting $f$ so that $f''(\tau)=0$ and $f''(\tau)f(\tau)=f'(\tau)^2$. This in turn allows access to all terms intervening in $[m_b]_a$ (as per \eqref{eq:thm:main mean}), making it possible to choose $f$ so to maximize the distances $|[m_1]_1-[m_1]_2|$ and $|[m_2]_2-[m_2]_1|$. However, in addition to the cumbersome aspect of the induced procedure (and the instability implied by multiple evaluations of the scores $F$ under several settings for $f$ and $c_{[l]a}$), such operations also alter the values of the variances in \eqref{eq:thm:main covariance} for which not all terms are easily estimated. It thus seems more delicate to derive a simple method to optimize $f$ in addition to $\alpha$.

\section{Concluding Remarks}
\label{sec:conclusion}

This article is part of a series of works consisting in evaluating the performance of kernel-based machine learning methods in the large dimensional data regime \citep{couillet2015kernel,liao2017large,kammoun2017}. Relying on the derivations in \citep{couillet2015kernel} that provide a Taylor expansion of radial kernel matrices around the limiting common value $\tau$ of $\frac1p\|x_i-x_j\|^2$ for $i\neq j$ and $p\to\infty$, we observed that the choice of the kernel function $f$ merely affects the classification performances through the successive derivatives of $f$ at $\tau$. In particular, similar to \citep{couillet2015kernel,liao2017large} (which then motivated the study \citep{kammoun2017}), we found that the case $f'(\tau)=0$ induces a sharp phase transition on normalized data by which the asymptotic classification error rate vanishes. However, unlike \citep{couillet2015kernel,liao2017large}, the exact expression at the core of the limiting performance assumes a different form. Of importance is the finding that, under a heat kernel assumption $f(t)=\exp(-\frac{t}{2\sigma^2})$, the studied semi-supervised learning method fails to classify Gaussian mixtures of the type $\mathcal N(0,C_k)$ with $\tr C_k^\circ=O(\sqrt{p})$ and $\tr C_kC_{k'}-\tr C_k^2=o(p)$, which unsupervised learning or LS-SVM are able to do \citep{couillet2015kernel,liao2017large}. This paradox may deserve a more structural way of considering together methods on the spectrum from unsupervised to supervised learning. 

\medskip

The very fact that the kernel matrix $W$ is essentially equivalent to the matrix $f(\tau)1_n1_n^\trans$ (the $n\times n$ matrix filled with $f(\tau)$ values), thereby strongly disrupting with the expected natural behavior of kernels, essentially follows from the Gaussian mixture model we assumed as well as from the decision to compare vectors by means of a mere Euclidean distance. We believe that this simplistic (although widely used) method explains the strong coincidence between performances on the Gaussian mixture model and on realistic datasets. Indeed, as radial functions are not specially adapted to image vectors (as would be wavelet or convolutional filters), the kernel likely operates on first order statistics of the input vectors, hence similar to its action on a Gaussian mixture dataset. It would be interesting to generalize our result, and for that matter the set of works \citep{couillet2015kernel,liao2017large,kammoun2017}, to more involved data-oriented kernels, so long that the data contain enough exploitable degrees of freedom.

\medskip

It is also quite instructive to note that, from the proof of our main results, the terms remaining after the expansion of $D_{[u]}^{-1-\alpha}W_{[uu]}D_{[u]}^\alpha$ appearing in \eqref{eq:opt-result} almost all vanish, strongly suggesting that similar results would be obtained if the inverse matrix in \eqref{eq:opt-result} were discarded altogether. This implies that the intra-unlabelled data kernel $W_{[uu]}$ is of virtually no asymptotic use. A promising avenue of investigation would consist in introducing appropriate scaling parameters in the label propagation method or the optimization problem \eqref{eq:opt} to ensure that $W_{[uu]}$ is effectively used in the algorithm. Early simulations do suggest that elementary amendments to \eqref{eq:opt-result} indeed result in possibly striking performance improvements.

These considerations are left to future works.

\acks{This work is supported by the ANR Project RMT4GRAPH (ANR-14-CE28-0006).}

\appendix

\section{Preliminaries}

We begin with some additional notations that will be useful in the proofs.
\begin{itemize}
	\item For $x_{i}\in\mathcal{C}_{k}$, $\omega_{i}\equiv  (x_{i}-\mu_{k})/\sqrt{p}$, and $\Omega\equiv [\omega_{1},\cdots,\omega_{n}]^\trans$
	\item $j_{k}\in\mathbb{R}^{n}$ is the canonical vector of $\mathcal{C}_{k}$, in the sense that its $i$-th element is $1$ if $x_{i}\in\mathcal{C}_{k}$ or $0$ otherwise. $j_{[l]k}$ and $j_{[u]k}$ are respectively the canonical vectors for labelled and unlabelled data of $\mathcal{C}_k$.
	\item $\psi_{i}\equiv\Vert \omega_{i} \Vert^{2}-{\rm E}[\Vert \omega_{i} \Vert^{2}]$, $\psi\equiv [\psi_{1},\cdots,\psi_{n}]^\trans$ and $(\psi)^{2}\equiv [(\psi_{1})^{2},\cdots,(\psi_{n})^{2}]^\trans$.
\end{itemize}

With these notations at hand, we introduce next the generalized version of Theorem~\ref{thm:main} for all $\alpha=O(1)$ (rather than $\alpha=-1+O(1/\sqrt{n})$).

\begin{theorem}
	\label{thm:behaviour of output Fu}
	For $x_i\in\mathcal C_b$ an unlabelled vector (i.e., $i>n_{[l]}$), let $\hat F_{ia}$ be given by \eqref{eq:normalized_scores} with $F$ defined in \eqref{eq:opt-result} for $\alpha=O(1)$. Then, under Assumptions~\ref{Growth-Rate}--\ref{Kernel-function},
	\begin{align}
		p\hat F_{i\cdot} &= p(1+z_i)1_K+ G_i + o_P(1) \label{eq:thm Fu eq1}\\
		G_i& \sim \mathcal N( m_b , \Sigma_b )\label{eq:thm Fu eq2}
	\end{align}
	where $z_i$ is as in Theorem~\ref{thm:main} and
	\begin{itemize}
		\item[(i)] for $F_{i\cdot}$ considered on the $\sigma$-field induced by the random variables $x_{[l]+1},\ldots,x_n$, $p=1,2,\ldots$,
			\begin{align}
				[m_{b}]_{a} &=H_{ab}+\frac{1}{n_{[l]}}\sum_{d=1}^{K}(\alpha n_{d}+n_{[u]d})H_{ad}+(1+\alpha)\frac{ n}{n_{[l]}}\Bigg[\Delta_{a}+\frac{p}{n_{[l]a}}\frac{f'(\tau)}{f(\tau)}\psi_{[l]}^Tj_{[l]a}-\alpha\frac{f'(\tau)^2}{f(\tau)^2}t_{a}t_{b}\Bigg]\label{eq:thm Fu eq7}\\
				[\Sigma_{b}]_{a_{1}a_{2}} &=\Bigg(\frac{(-\alpha^{2}-\alpha)n-n_{[l]}}{n_{[l]}}\frac{f'(\tau)^{2}}{f(\tau)^{2}}+\frac{f''(\tau)}{f(\tau)}\Bigg)^{2}T_{bb}t_{a_{1}}t_{a_{2}}+{\bm\delta}_{a_1}^{a_2}\frac{f'(\tau)^{2}}{f(\tau)^{2}}\frac{4c_0T_{ba_1}}{c_{[l]a_1}}+\frac{4f'(\tau)^{2}}{f(\tau)^{2}}\mu^{\circ}_{a_{1}}C_{b}\mu^{\circ}_{a_{2}}\label{eq:thm Fu eq8}
			\end{align}
			where
			\begin{align}
				H_{ab}&=\frac{f'(\tau)}{f(\tau)}\lVert\mu^{\circ}_{b}-\mu^{\circ}_{a}\rVert^{2}+\left( \frac{f''(\tau)}{f(\tau)} - \frac{f'(\tau)^2}{f(\tau)^2} \right)t_{a}t_{b}+\frac{2f''(\tau)}{f(\tau)}T_{ab}\label{eq:thm Fu eq5}\\
				\Delta_{a}&=\frac{\sqrt{p}f'(\tau)}{f(\tau)}t_{a}+\frac{\alpha f'(\tau)^{2}+f(\tau)f''(\tau)}{2f(\tau)^{2}}\left(2T_{aa}+t_{a}^{2}\right)+\frac{1}{n_{[l]}}\left(\frac{f'(\tau)}{f(\tau)}\right)^{2}\left(\sum_{d=1}^{K}n_{[u]d}t_{d}\right)t_{a}\label{eq:thm Fu eq6}.
			\end{align} 

		\item[(ii)] for $F_{i\cdot}$ considered on the $\sigma$-field induced by the random variables $x_1,\ldots,x_n$,  

			\begin{align}
				[m_{b}]_{a} & =H_{ab}+\frac{1}{n_{[l]}}\sum_{d=1}^{K}(\alpha n_{d}+n_{[u]d})H_{ad}+(1+\alpha)\frac{ n}{n_{[l]}}\Bigg[\Delta_{a}-\alpha\frac{f'(\tau)^2}{f(\tau)^2}t_{a}t_{b}\Bigg]\label{eq:thm Fu eq3}\\
				[\Sigma_{b}]_{a_{1}a_{2}} & =\Bigg(\frac{(-\alpha^{2}-\alpha)n-n_{[l]}}{n_{[l]}}\frac{f'(\tau)^{2}}{f(\tau)^{2}}+\frac{f''(\tau)}{f(\tau)}\Bigg)^{2}T_{bb}t_{a_{1}}t_{a_{2}} \nonumber \\
				&+{\bm\delta}_{a_1}^{a_2}\frac{f'(\tau)^{2}}{f(\tau)^{2}}\Bigg(\frac{(1+\alpha)^{2}}{c_{[l]}^{2}}\frac{2c_0T_{aa}}{c_{[l]a_1}}+\frac{4c_0T_{ba_1}}{c_{[l]a_1}}\Bigg) +\frac{4f'(\tau)^{2}}{f(\tau)^{2}}\mu^{\circ}_{a_{1}}C_{b}\mu^{\circ}_{a_{2}}\label{eq:thm Fu eq4}
			\end{align}
			with $H_{ab}$ given in \eqref{eq:thm Fu eq5} and $\Delta_{a}$ in \eqref{eq:thm Fu eq6}.
	\end{itemize}

\end{theorem}

Let $P(x_{i}\to\mathcal{C}_{b}|x_{i}\in\mathcal{C}_{b},x_{1},\cdots,x_{n_{[l]}})$ denote the probability of correct classification of $x_i\in\mathcal{C}_{b}$ unlabelled, conditioned on $x_1,\ldots,x_{n_{[l]}}$, and $P(x_{i}\to\mathcal{C}_{b}|x_{i}\in\mathcal{C}_{b})$ the unconditional probability. Recall that the probability of correct classification of $x_i\in\mathcal{C}_{b}$ is the same as the probability of $\hat{F}_{ib}>\max_{a\neq b}\hat{F}_{ib}$, which, according to the above theorem, is asymptotically the probability that $[G_i]_b$ is the greatest element of $G_i$. Particularly for $K=2$, we have the following corollary.

\begin{corollary}
	\label{cor:conditional precision}
	Under the conditions of Theorem 1, and with $K=2$, we have, for $a\neq b\in\{1,2\}$,
	\begin{itemize}
		\item[(i)] Conditionally on $x_{1},\cdots,x_{n_{[l]}}$,
			\begin{align}\label{eq:conditional precision}
				&\mathbb{P}\left(x_{i}\to\mathcal{C}_{b}|x_{i}\in\mathcal{C}_{b},x_{1},\cdots,x_{n_{[l]}})-\Phi({\theta}_{b}^{a}\right)\to 0 \\
				&{\theta}_{b}^{a}=\frac{[m_{b}]_{b}-[m_{b}]_{a}}{\sqrt{[\Sigma_{b}]_{bb}+[\Sigma_{b}]_{aa}-2[\Sigma_{b}]_{ab}}} \nonumber
			\end{align}
			where  $\Phi(u)=\frac{1}{2\pi}\int_{-\infty}^{u}\exp(-t^{2}/2)dt$ and $m_{b}$, $\Sigma_{b}$ are given in (i) of Theorem~\ref{thm:behaviour of output Fu}.
	
		\item[(ii)] Unconditionally,
			\begin{align}\label{eq:unconditional precision}
				&\mathbb{P}\left(x_{i}\to\mathcal{C}_{b}|x_{i}\in\mathcal{C}_{b})-\Phi({\theta}_{b}^{a}\right)\to 0 \\
				&{\theta}_{b}^{a}=\frac{[m_{b}]_{b}-[m_{b}]_{a}}{\sqrt{[\Sigma_{b}]_{bb}+[\Sigma_{b}]_{aa}-2[\Sigma_{b}]_{ab}}} \nonumber
			\end{align}
			where here $m_{b}$, $\Sigma_{b}$ are given in (ii) of Theorem~\ref{thm:behaviour of output Fu}.
	\end{itemize}	
\end{corollary}

The remainder of the appendix is dedicated to the proof of Theorem~\ref{thm:behaviour of output Fu} and Corollary~\ref{cor:conditional precision} from which the results of Section~\ref{sec:results} directly unfold.

\section{Proof of Theorems~\ref{thm:behaviour of output Fu}}
\label{sec:proof}
The proof of Theorem~\ref{thm:behaviour of output Fu} is divided into two steps: first, we Taylor-expand the normalized scores for unlabelled data $\hat{F}_{[u]}$ using the convergence $\frac1p\|x_i-x_j\|^2\asto \tau$ for all $i\neq j$; this expansion yields a random equivalent $\hat{F}_{[u]}^{\text{eq}}$ in the sense that $p(\hat{F}_{[u]}-\hat{F}_{[u]}^{\text{eq}})\asto 0$. Proposition~\ref{prop:first-orders-of-Fu} is directly obtained from $\hat{F}_{[u]}^{\text{eq}}$. We then complete the proof by demonstrating the convergence to Gaussian variables of $\hat{F}_{[u]}^{\text{eq}}$ by means of a central limit theorem argument.

\subsection{Step 1: Taylor expansion}
In the following, we provide a sketch of the development of $F_{[u]}$; most unshown intermediary steps can be retrieved from simple, yet painstaking algebraic calculus.

Recall from \eqref{eq:opt-result} the expression of the unnormalized scores for unlabelled data
\begin{align*}
	F_{[u]}=(I_{n_{u}}-D_{[u]}^{-1-\alpha}W_{[uu]}D_{[u]}^{\alpha})^{-1}D_{[u]}^{-1-\alpha}W_{[ul]}D_{[l]}^{\alpha}F_{[l]}.
\end{align*}
We first proceed to the development of the terms $W_{[ul]}$, $W_{[uu]}$, subsequently to $D_{[l]},D_{[u]}$, to then reach an expression for $F_{[u]}$. To this end, owing to the convergence $\|x_i-x_j\|^2/p\asto \tau$ for all $i\neq j$, we first Taylor-expand $W_{ij}=f(\| x_{i}-x_{j}\|^2/p)$ around $f(\tau)$ to obtain the following expansion for $W$, already evaluated in \citep{couillet2015kernel},
\begin{equation}
	\label{eq:W}
	W = W^{(n)}+W^{(\sqrt{n})}+W^{(1)}+O(n^{-\frac{1}{2}})
\end{equation}
where $\|W^{(n)}\|=O(n)$, $\|W^{(\sqrt{n})}=O(\sqrt{n})$ and $\|W^{(1)}\|=O(1)$, with the definitions
\begin{align}
	W^{(n)}&=\ftau1_{n}1_{n}^\trans \\
	\label{eq:W2}
	W^{(\sqrt{n})} &=\fftau\left[\psi 1_{n}^\trans+1_{n}\psi^\trans+\left(\sum_{b=1}^{K}\frac{t_{b}}{\sqrt{p}}j_{b}\right)1_{n}^\trans+1_{n}\sum_{a=1}^{K}\frac{t_{a}}{\sqrt{p}}j_{a}^\trans\right]\\
	\label{eq:W3}
	W^{(1)}&=\fftau\Bigg[\sum_{a,b=1}^{K}\frac{\|\muctra-\muctrb\|^2}{p}j_{b}j_{a}^\trans-\frac{2}{\sqrt{p}}\Omega \sum_{a=1}^{K}\muctra j_{a}^\trans+\frac{2}{\sqrt{p}}\sum_{b=1}^{K}{\rm diag}(j_{b})\Omega \muctrb 1_n^\trans-\frac{2}{\sqrt{p}}\sum_{b=1}^{K}j_{b}\mu_{b}^{\circ \trans}\Omega^\trans\nonumber\\
	&+\frac{2}{\sqrt{p}}1_{n}\sum_{a=1}^{K}{\muctra}^\trans\Omega^\trans {\rm diag}(j_{a})-2\Omega\Omega^\trans\Bigg]+\frac{\ffftau}{2}\bigg[(\psi)^21_{n}^\trans+1_{n}[(\psi)^2]^\trans+\sum_{b=1}^{K}\frac{t_b^2}{p}j_{b} 1_n^\trans+1_{n}\sum_{a=1}^{K}\frac{t_a^2}{p}j_{a}^\trans \nonumber\\
	&+2\sum_{a,b=1}^{K}\frac{t_{a}t_{b}}{p}j_{b}j_{a}^\trans+2\sum_{b=1}^{K}{\rm diag}(j_{b})\frac{t_{b}}{\sqrt{p}}\psi 1_{n}^\trans+2\sum_{b=1}^{K}\frac{t_{b}}{\sqrt{p}}j_{b}\psi^\trans+2\sum_{a=1}^{K}1_{n}\psi^\trans {\rm diag}(j_{a})\frac{t_{a}}{\sqrt{p}}+2\psi \sum_{a=1}^{K}\frac{t_{a}}{\sqrt{p}}j_{a}^\trans\nonumber\\
&+4\sum_{a,b=1}^{K}\frac{T_{ab}}{p}j_{b}j_{a}^\trans+2\psi\psi^\trans \bigg]+(f(0)-\ftau+\tau\fftau)I_{n}.
	\end{align}

	As $W_{[ul]}$, $W_{[uu]}$ are sub-matrices of $W$, their approximated expressions are obtained directly by extracting the corresponding subsets of \eqref{eq:W}. Applying then \eqref{eq:W} in $D={\rm diag}(W1_{n})$, we next find
	\begin{align}
		\label{eq:D}
		D=nf(\tau)\bigg[I_{n}+\frac{1}{n\ftau}{\rm diag}(W^{(\sqrt{n})}1_{n}+W^{(1)}1_{n})\bigg]+O(n^{-\frac{1}{2}}).
	\end{align}
Thus, for any $\sigma\in\mathbb{R}$, $(n^{-1}D)^{\sigma}$ can be Taylor-expanded around $\ftau^{\sigma}I_{n}$ as
\begin{align}
\label{eq:Taylor-expanded D}
(n^{-1}D)^{\sigma}&=\ftau^{\sigma}\bigg[I_{n}+\frac{\sigma1}{n\ftau}{\rm diag}((W^{(\sqrt{n})}+W^{(1)})1_{n}) + \frac{\sigma(\sigma-1)}{2n^{2}\ftau^{2}}{\rm diag}^{2}(W^{(\sqrt{n})}1_{n})\bigg]+O(n^{-\frac{3}{2}})
\end{align}
where ${\rm diag}^{2}(\cdot)$ stands for the squared diagonal matrix. The Taylor-expansions of $(n^{-1}D_{[u]})^{\alpha}$ and $(n^{-1}D_{[l]})^{\alpha}$ are then directly extracted from this expression for $\sigma=\alpha$, and similarly for $(n^{-1}D_{[u]})^{-1-\alpha}$ with $\sigma=-1-\alpha$. Since 
\begin{align*}
	D_{[u]}^{-1-\alpha}W_{[ul]}D_{[l]}^{\alpha}=\frac{1}{n}(n^{-1}D_{[u]})^{-1-\alpha}W_{[ul]}(n^{-1}D_{[l]})^{\alpha}
\end{align*}
it then suffices to multiply the Taylor-expansions of $(n^{-1}D_{[u]})^{\alpha}$, $(n^{-1}D_{[l]})^{\alpha}$, and $W_{[ul]}$, given respectively in \eqref{eq:Taylor-expanded D} and \eqref{eq:W}, normalize by $n$ and then organize the result in terms of order $O(1)$, $O(1/\sqrt{n})$, and $O(1/n)$. 

The term $D_{[u]}^{-1-\alpha}W_{[uu]}D_{[u]}^{\alpha}$ is dealt with in the same way. In particular
\begin{equation}
	\label{eq:Luu}
	D_{[u]}^{-1-\alpha}W_{[uu]}D_{[u]}^{\alpha}=\frac{1}{n}1_{n_{[u]}}1_{n_{[l]}}+O(n^{-\frac{1}{2}}).
\end{equation}
Therefore, $(I_{n_{[u]}}-D_{[u]}^{-1-\alpha}W_{[uu]}D_{[u]}^{\alpha})^{-1}$ may be simply written as
\begin{equation}
	\left(I_{n_{[u]}}-\frac{1}{n}1_{n_{[u]}}1_{n_{[u]}} + O(n^{-\frac{1}{2}}) \right)^{-1}=I_{n_{[u]}}+\frac{1}{n_{[l]}}1_{n_{[u]}}1_{n_{[u]}} + O(n^{-\frac{1}{2}}).
\end{equation}
Combining all terms together completes the full linearization of $\hat{F}_{[u]}$. 

This last derivation, which we do not provide in full here, is simpler than it appears and is in fact quite instructive in the overall behavior of $F^{[u]}$. Indeed, only product terms in the development of $(I_{n_{[u]}}-D_{[u]}^{-1-\alpha}W_{[uu]}D_{[u]}^{\alpha})^{-1}$ and $D_{[u]}^{-1-\alpha}W_{[ul]}D_{[l]}^{\alpha}F^{[l]}$ of order at least $O(1)$ shall remain, which discards already a few terms. Now, in addition, note that for any vector $v$, $v1_{n_{[l]}}^\trans F^{[l]}=v1_k^\trans$ so that such matrices are non informative for classification (they have identical score columns); these terms are all placed in the intermediary variable $z$, the entries $z_i$ of which are irrelevant and thus left as is (these are the $z_i$'s of Proposition~\ref{prop:first-orders-of-Fu} and Theorem~\ref{thm:main}). It is in particular noteworthy to see that {\it all} terms of $W_{[uu]}^{(1)}$ that remain after taking the product with $D_{[u]}^{-1-\alpha}W_{[ul]}D_{[l]}^{\alpha}F^{[l]}$ are precisely those multiplied by $f(\tau)1_{n_{[u]}}1_{n_{[l]}}^\trans F^{[l]}$ and thus become part of the vector $z$. Since most informative terms in the kernel matrix development are found in $W^{(1)}$, this means that the algorithm under study shall make little use of the {\it unsupervised} information about the data (those found in $W_{[uu]}^{(1)}$). This is an important remark which, as discussed in Section~\ref{sec:conclusion}, opens up the path to further improvements of the semi-supervised learning algorithms which would use more efficiently the information in $W_{[uu]}^{(1)}$.

All calculus made, this development finally leads to $F_{[u]}=F_{[u]}^{\text{eq}}$ with, for $a,b\in\{1,\ldots,K\}$ and $x_i\in\mathcal C_b$, $i>n_{[l]}$,
\begin{align}
	\label{eq:developed_normalized_scores}
	\hat{F}_{ia}^{\text{eq}} &=1+\frac{1}{p}\Bigg[H_{ab}+\frac{1}{n_{[l]}}\sum_{d=1}^{K}H_{ad}(\alpha n_{d}+n_{[u]d})\Bigg]+(1+\alpha)\frac{ n}{pn_{[l]}}\Bigg[\Delta_{a}-\alpha\frac{f'(\tau)^2}{f(\tau)^2}t_{a}t_{b}\Bigg]+\frac{2f'(\tau)}{f(\tau)\sqrt{p}}\mu^{\circ}_{a}\omega_{i}\nonumber\\
	&+\Bigg(\frac{(-\alpha^{2}-\alpha)n-n_{[l]}}{n_{[l]}}\frac{f'(\tau)^{2}}{f(\tau)^{2}}+\frac{\ffftau}{f(\tau)}\Bigg)\frac{t_{a}}{\sqrt{p}}\psi_{i}+\frac{f'(\tau)}{f(\tau)}\Bigg(\frac{(1+\alpha)n}{n_{[l]}n_{[l]a}}\psi_{[l]}^\trans j_{[l]a}+\frac{4}{n_{[l]a}}j_{[l]a}^T\Omega_{[l]}\omega_{i}\Bigg)+z_{i}
\end{align}
where $H_{ab}$ is as specified in \eqref{eq:thm Fu eq5}, $\Delta_{a}$ as in \eqref{eq:thm Fu eq6}, and $z_i=O(\sqrt{p})$ is some residual random variable only dependent on $x_i$. Gathering the terms in successive orders of magnitude, Proposition~\ref{prop:first-orders-of-Fu} is then straightforwardly proven from \eqref{eq:developed_normalized_scores}.

\subsection{Step 2: Central limit theorem}
The focus of this step is to examine $\mathcal{G}_i=p(\hat{F}_{i}^{\text{eq}}-(1+z_i)1_K)$. Theorem~\ref{thm:behaviour of output Fu} can be proven by showing that $\mathcal{G}_i=G_i+o_P(1)$.

First consider Item (i) of Theorem~\ref{thm:behaviour of output Fu}, which describes the behavior of $\hat{F}_{[u]}$ conditioned on $x_1,\ldots,x_{n_{[l]}}$. Recall that a necessary and sufficient condition for a vector $v$ to be a Gaussian vector is that all linear combinations of the elements of $v$ are Gaussian variables. Thus, for given $x_1,\ldots,x_{n_{[l]}}$ deterministic, according to \eqref{eq:developed_normalized_scores}, $\mathcal{G}_i$ is asymptotically Gaussian if, for all $g_1\in\RR$, $g_2\in\RR^{p}$, $g_1\psi_{i}+g_2^T\omega_{i}$ has a central limit. 

Letting $\omega_{i}=\frac{C_b^{\frac{1}{2}}}{\sqrt{p}}r$, with $r\sim \mathcal N( 0, I_p )$, $g_1\psi_{i}+g_2\omega_{i}$ can be rewritten as $r^\trans Ar+br+c$ with $A=g_1\frac{C_b}{p}$, $b=g_2\frac{C_b}{p}$, $c=-g_1\frac{{\rm tr} C_b}{p}$. Since $A$ is symmetric, there exists an orthonormal matrix $U$ and a diagonal $\Lambda$ such that $A=U^\trans\Lambda U$. We thus get 
\begin{align*}
	r^\trans Ar+br+c=r^TU^T\Lambda Ur+bU^TUr+c=\tilde{r}^T\Lambda\tilde{r}+\tilde{b}\tilde{r}+c
\end{align*}
with $\tilde{r}=Ur$ and $\tilde{b}=bU^\trans$. By unitary invariance, we have $\tilde{r}\sim \mathcal N( 0, I_p )$ so that $g_1\psi_{i}+g_2\omega_{i}$ is thus the sum of the independent but not identically distributed random variables $q_j=\lambda_j\tilde{r}_j^2+\tilde{b}_j\tilde{r}_j$, $i=1,\ldots,p$. From Lyapunov's central limit theorem \citep[Theorem~27.3]{billingsley2008probability}, it remains to find a $\delta>0$ such that $\frac{\sum_j \mathbb{E}|q_j-\mathbb{E}[q_j]|^{2+\delta}}{\left(\sum_{j}{\rm Var}[q_j]\right)^{1+\delta/2}}\to0$ to ensure the central limit theorem. For $\delta=1$, we have $\mathbb{E}[q_j]=\lambda_j$, ${\rm Var}[q_j]=2\lambda_j^2+\tilde{b}_j^2$ and $\mathbb{E}\left[(q_j-\mathbb{E}[q_j])^3\right]=8\lambda_j^3+6\lambda_j\tilde{b}_j^2$, so that $\frac{\sum_j \mathbb{E}\left[|q_j-\mathbb{E}[q_j]|^{3}\right]}{\left(\sum_{j}{\rm Var}[q_j]\right)^{3/2}}=O(n^{-\frac{1}{2}})$.

It thus remains to evaluate the expectation and covariance matrix of $\mathcal{G}_i$ conditioned on $x_1,\ldots,x_{n_{[l]}}$ to obtain (i) of Theorem~\ref{thm:behaviour of output Fu}. For $x_i\in\mathcal{C}_b$, we have
\begin{align*}
	\mathbb{E}\{[\mathcal{G}_i]_a\}&=H_{ab}+\frac{1}{n_{[l]}}\sum_{d=1}^{K}(\alpha n_{d}+n_{[u]d})H_{ad}\nonumber \\
	&+(1+\alpha)\frac{ n}{n_{[l]}}\Bigg[\Delta_{a}+\frac{p}{n_{[l]a}}\frac{f'(\tau)}{f(\tau)}\psi_{[l]}^Tj_{[l]a}-\alpha\frac{f'(\tau)^2}{f(\tau)^2}t_{a}t_{b}\Bigg]\\
	{\rm Cov}\{[\mathcal{G}_i]_{a_1}[\mathcal{G}_i]_{a_2}\}&=\Bigg(\frac{(-\alpha^{2}-\alpha)n-n_{[l]}}{n_{[l]}}\frac{f'(\tau)^{2}}{f(\tau)^{2}}+\frac{f''(\tau)}{f(\tau)}\Bigg)^{2}T_{bb}t_{a_{1}}t_{a_{2}}\nonumber \\
	&+{\bm\delta}_{a_{1}}^{a_{2}}\frac{f'(\tau)^{2}}{f(\tau)^{2}}\frac{4c_0T_{ba_1}}{c_{[l]a_1}}+\frac{4f'(\tau)^{2}}{f(\tau)^{2}}\mu^{\circ}_{a_{1}}C_{b}\mu^{\circ}_{a_{2}}+o(1).
\end{align*}
From the above equations, we retrieve the asymptotic expressions of $[m_{b}]_{a}$ and $[\Delta_b]_{a_1a_2}$ given in \eqref{eq:thm Fu eq7} and \eqref{eq:thm Fu eq8}. This completes the proof of Item (i) of Theorem~\ref{thm:behaviour of output Fu}. Item (ii) is easily proved by following the same reasoning.

\bibliography{RMT4SSL}

\end{document}